\documentclass[runningheads]{llncs}
\usepackage{graphicx}
\usepackage{epsfig}
\usepackage{comment}
\usepackage{mathtools}
\usepackage{amsmath,amssymb} 
\usepackage{color}
\usepackage{mynotation}
\usepackage{bbm}
\usepackage{algorithm}
\usepackage[noend]{algpseudocode}
\usepackage{booktabs}
\usepackage[caption=false]{subfig}
\usepackage{pifont}
\usepackage{arydshln}
\usepackage[width=122mm,left=12mm,paperwidth=146mm,height=193mm,top=12mm,paperheight=217mm]{geometry}
\usepackage{verbatim}

\newcommand{\parsection}[1]{\noindent\textbf{#1:} }

\newcommand{\corr}{p}
\newcommand{\corrat}[2]{\corr(\vecn{#1}|\vecn{#2})}

\newcommand{\cv}{\mathbf{CV}}
\newcommand{\cvat}[2]{\cv(\vecn{#1}, \vecn{#2})}

\newcommand{\init}{\Upsilon}
\newcommand{\prop}{\Pi}
\newcommand{\propconf}{\xi}
\newcommand{\predictor}{P}
\newcommand{\stateupdate}{\Phi}
\newcommand{\fusedscore}{\varsigma}

\newcommand{\motionfeat}{x}
\newcommand{\eg}{e.g.}
\newcommand{\ie}{i.e.}
\newcommand{\etal}{et al}

\algrenewcommand{\algorithmicrequire}{\textbf{Input:}}

\begin{document}
\pagestyle{headings}
\mainmatter

\title{Know Your Surroundings: Exploiting Scene Information for Object Tracking} 


\titlerunning{Know Your Surroundings: Exploiting Scene Information for Object Tracking} 
\authorrunning{Goutam Bhat, Martin Danelljan, Luc Van Gool, and Radu Timofte} 
\newcommand{\aand}{\hspace{6mm}}
\author{Goutam Bhat  \aand  Martin Danelljan \aand Luc Van Gool \aand Radu Timofte
}
\institute{Computer Vision Lab, ETH Z\"urich, Switzerland}

\maketitle
\begin{abstract}
Current state-of-the-art trackers only rely on a target appearance model in order to localize the object in each frame. Such approaches are however prone to fail in case of \eg fast appearance changes or presence of distractor objects, where a target appearance model alone is insufficient for robust tracking. Having the knowledge about the presence and locations of other objects in the surrounding scene can be highly beneficial in such cases. This scene information can be propagated through the sequence and used to, for instance, explicitly avoid distractor objects and eliminate target candidate regions.

In this work, we propose a novel tracking architecture which can utilize scene information for tracking. Our tracker represents such information as dense localized state vectors, which can encode, for example, if the local region is target, background, or distractor. These state vectors are propagated through the sequence and combined with the appearance model output to localize the target. Our network is learned to effectively utilize the scene information by directly maximizing tracking performance on video segments. The proposed approach sets a new state-of-the-art on $3$ tracking benchmarks, achieving an AO score of $63.6\%$ on the recent GOT-10k dataset.
\end{abstract}

\section{Introduction}
\begin{figure}[t]
	\centering%
	\newcommand{\wid}{0.6\columnwidth}%
	\includegraphics*[trim = 0 0 0 0, width = \wid]{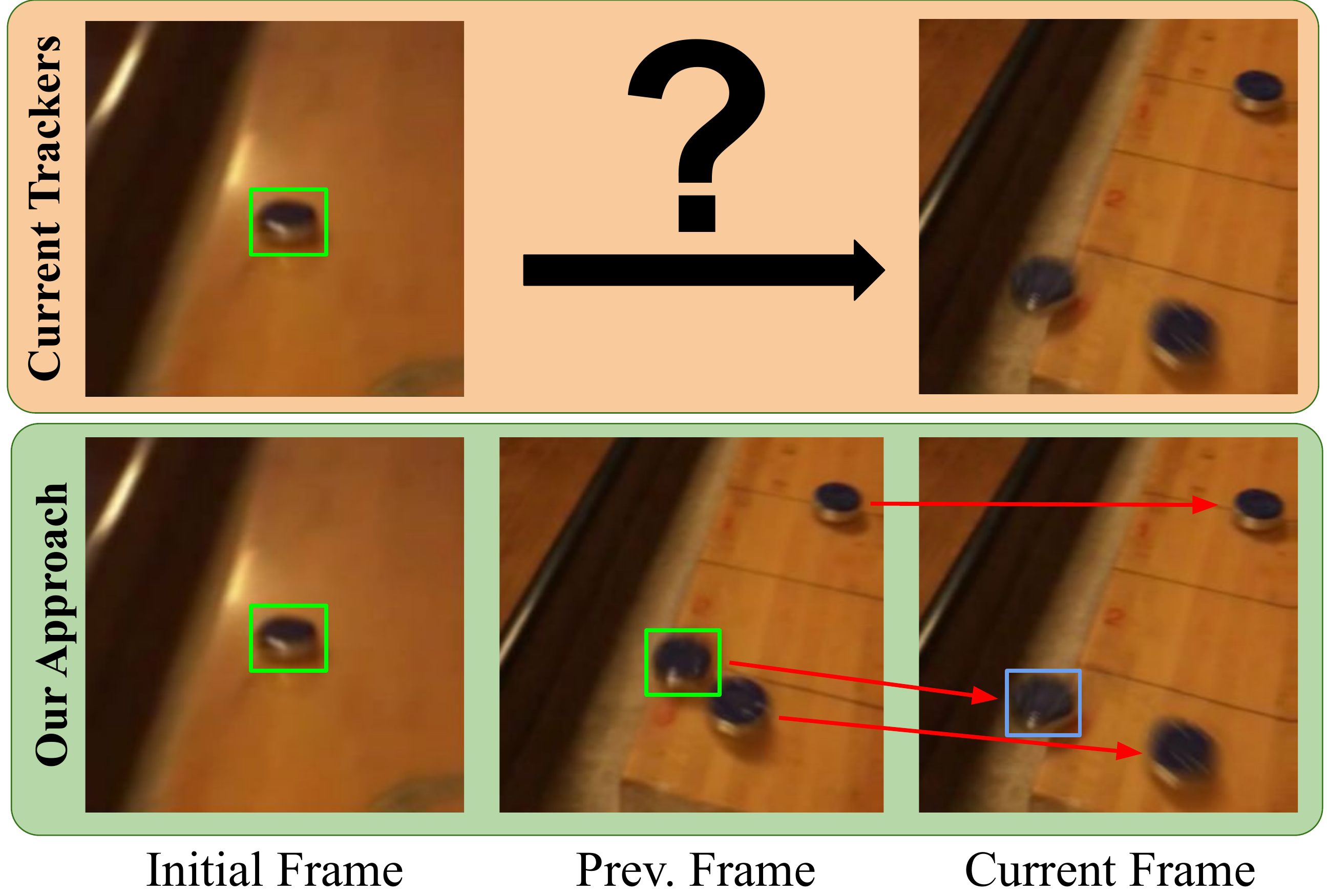}\vspace{-4mm}%
	\caption{Current approaches (top) only utilize an appearance model to track the target object. However, such a  strategy fails in the above example. Here, the presence of distractor objects  makes it virtually impossible to correctly localize the target based on appearance only, even if the appearance model is continuously updated using previous frames. In contrast, our approach (bottom) is also aware of other objects in the scene. This scene information is propagated through the sequence by computing a dense correspondence (red arrows) between consecutive frames. The propagated scene knowledge greatly simplifies the target localization problem, allowing us to reliably track the target.}\vspace{-6.0mm}%
	\label{fig:intro}%
\end{figure}

Generic object tracking is one of the fundamental computer vision problems with numerous applications. The task is to estimate the state of a target object in each frame of a video sequence, given only its initial appearance. Most current approaches~\cite{DiMP,ATOM,SiamRPN++,TADT,ASRCF,GCT,MDNet} tackle the problem by learning an appearance model of the target in the initial frame. This model is then applied in subsequent frames to localize the target by distinguishing its appearance from the surrounding background.  
While achieving impressive tracking performance~\cite{VOT2018,GOT10k}, these approaches rely \emph{only} on the appearance model, and do not utilize any other information contained in the scene. 

In contrast, humans exploit a much richer set of cues when tracking an object. We have a holistic view of the scene, taking into consideration not only the target object, but are also continuously aware of the other objects in the scene. Such information is helpful when localizing the target, \eg\ in case of cluttered scenes with distractor objects, or when the target undergoes fast appearance change. Consider the example in Figure~\ref{fig:intro}. Given only the initial target appearance, it is hard to confidently locate the target due to the presence of distractor objects. However, if we also utilize the previous frame, we can easily detect the presence of distractors. This knowledge can then be propagated to the next frame in order to reliably localize the target. While existing approaches update the appearance model with previously tracked frames, such a strategy by itself cannot capture the locations and characteristics of the other objects in the scene. 

In this work, we aim to go beyond the conventional frame-by-frame detection-based tracking. We propose a novel tracking architecture which can propagate valuable scene information through the sequence. This information is used to achieve an improved \emph{scene aware} target prediction in each frame. The scene information is represented using a dense set of localized state vectors. These encode valuable information about the local region, \eg\ whether the region corresponds to the target, background or a distractor object. As the regions move through the sequence, we propagate the corresponding state vectors by utilizing dense correspondence maps between frames. Consequently, our tracker is `aware' of every object in the scene and can use this information in order to \eg\ avoid distractor objects. This scene knowledge, along with the target appearance model, is used to predict the target state in each frame. The scene information captured by the state representation is then updated using a recurrent neural network module.

\newcommand{\bp}[1]{\textbf{#1}}

\parsection{Contributions}
Our main contributions are summarized as follows. 
\bp{(i)} We propose a novel tracking architecture that exploits rich scene information, represented as dense localized state vectors.
\bp{(ii)} A propagation module is introduced to map the states to subsequent frames by predicting soft correspondences.
\bp{(iii)}	 We develop a predictor module which effectively combines the output of the target appearance model with the propagated scene information in order to determine the target location.
\bp{(iv)} The states are updated with the new information by integrating a recurrent neural network module.
\bp{(v)} We train our network to directly maximize tracking performance on complete video segments.
	 
We perform comprehensive experiments on 5 challenging benchmarks: VOT-2018~\cite{VOT2018}, GOT-10k~\cite{GOT10k}, TrackingNet~\cite{TrackingNet}, OTB-100~\cite{OTB2015}, and NFS~\cite{NfS}. Our approach achieves state-of-the-art results on all five datasets. On the challenging GOT-10k dataset, our tracker obtains an average overlap (AO) score of $63.6\%$, outperforming the previous best approach by $2.5\%$. We also provide an ablation study analyzing the impact of key components in our tracking architecture.
\section{Related Work}
Most tracking approaches tackle the problem by learning an appearance model of the target in the first frame. A popular method to learn the target appearance model is the discriminative correlation filters (DCF) \cite{MOSSE2010,Henriques14,DanelljanICCV2015,DanelljanCVPR2017,BACFgaloogahi,STRCF}. These approaches exploit the convolution theorem to efficiently train a classifier in the Fourier domain using the circular shifts of the input image as training data. Another approach is to train or fine-tune a few layers of a deep neural network in the first frame to perform target-background classification~\cite{MDNet,ATOM,DiMP,Song2017CREST}. MDNet~\cite{MDNet} fine-tunes three fully-connected layers online, while DiMP \cite{DiMP} employs a meta-learning formulation to predict the weights of the classification layer. In recent years, Siamese networks have received significant attention \cite{SiameseFC,SiamRPN,SiamRPN++,RASNet,SASiamR}. These approaches address the tracking problem by learning a similarity measure, which is then used to locate the target.

The discriminative approaches discussed above exploit the background information in the scene to learn the target appearance model. Moreover, a number of attempts have been made to integrate background information into the appearance model in Siamese trackers~\cite{DaSiamRPN,Lee2019BilinearSN,Zhang_2019_ICCV}. However, in many cases, the distractor object is indistinguishable from a previous target appearance. Thus, a single target model is insufficient to achieve robust tracking in such cases. Further, in case of fast motion, it is hard to adapt the target model quickly to new distractors. In contrast to these works, our approach explicitly encodes localized information about different image regions and propagates this information through the sequence via dense matching. More related to our work,  \cite{Xiao2016DistractorSupportedST} aims to exploit the locations of distractors in the scene. However, it employs hand-crafted rules to classify image regions into background and target candidates independently in each frame, and employs a linear motion model to obtain final prediction. In contrast, we present a fully \emph{learnable} solution, where the encoding of image regions is learned and propagated  by appearance-based dense tracking between frames. Further, our final prediction is obtained combining the explicit background representation with the appearance model output.

In addition to appearance cues, a few approaches have investigated the use of optical flow information for tracking. Gladh~\etal~\cite{DeepMotionFeatureICPR} utilize deep motion features extracted from optical flow images to complement the appearance features when constructing the target model. Zhu~\etal~\cite{FlowTrack} use optical flow to warp the feature maps from the previous frames to a reference frame and aggregate them in order to learn the target appearance model. However, both these approaches utilize optical flow to only improve the robustness of the target model. In contrast, we explicitly use dense motion information to propagate information about background objects and structures in order to complement the target model. 

Some works have also investigated using recurrent neural networks (RNN) for object tracking. Gan~\etal~\cite{Gan2015FirstST} use a RNN to directly regress the target location using image features and previous target locations. Ning~\etal~\cite{2016SpatiallySR} utilize the YOLO~\cite{YOLO} detector to generate initial object proposals. These proposals, along with the image features, are passed through an LSTM~\cite{LSTM} to obtain the target box. Yang~\etal~\cite{Yang2017RecurrentFL,Yang2018LearningDM} use an LSTM to update the target model to account for changes in target appearance through a sequence. 
\section{Proposed Method}
We develop a novel tracking architecture capable of exploiting the scene information to improve tracking performance. While current state-of-the-art methods~\cite{ATOM,DiMP,SiamRPN++} only rely on the target appearance model to process every frame independently, our approach also propagates information about the scene from previous frames. This provides rich cues about the environment, \eg\ the location of distractor objects, which greatly aids the localization of the target. 

\begin{figure*}[t]
	\centering%
	\newcommand{\wid}{\textwidth}%
	\includegraphics*[trim = 0 160 150 0, width = \wid]{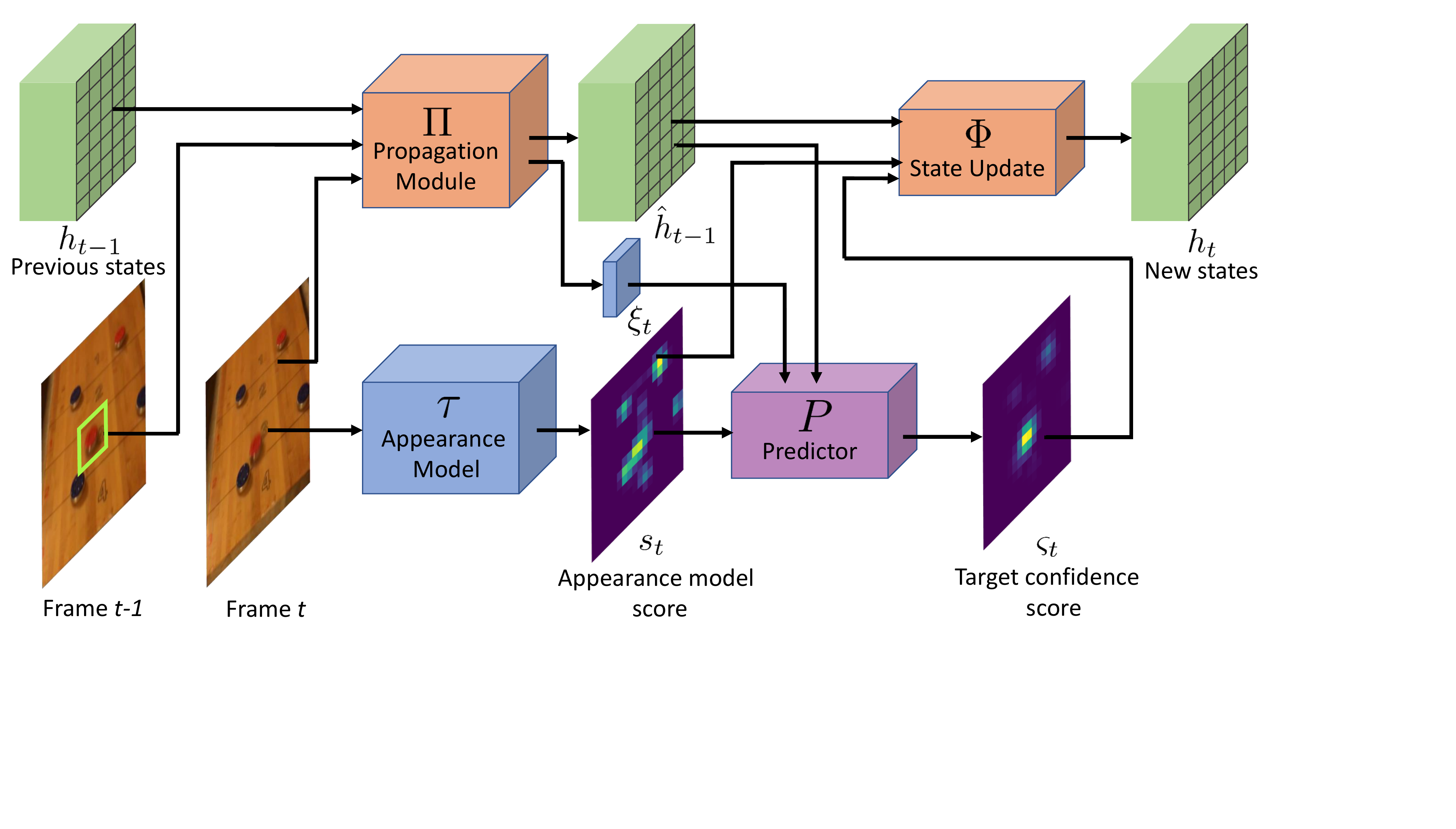}\vspace{-4mm}%
	\caption{An overview of our tracking architecture. In addition to using a target appearance model $\tau$, our tracker also exploits propagated scene information in order to track the target. The information about each image region is encoded within localized states $h$. Given the states $h_{t-1}$ from the previous frame, the propagation module $\prop$ maps these states from the previous frame to the current frame locations. These propagated states $\hat{h}_{t-1}$, along with the propagation reliability $\propconf_{t}$ and appearance model score $s_t$ are used by the predictor $\predictor$ to output the final target confidence scores $\fusedscore_t$. The state update module $\stateupdate$ then uses the current frame predictions to provide the new states $h_t$.
	}\vspace{-4mm}%
	\label{fig:network}%
\end{figure*}

A visual overview of our tracking architecture is provided in Figure~\ref{fig:network}. Our tracker internally tracks \emph{all} regions in the scene, and propagates any information about them that helps localization of the target. This is achieved by maintaining a state vector for every region in the target neighborhood. 
The state vector can, for instance, encode whether a particular patch corresponds to the target, background, or a distractor object that is likely to fool the target appearance model. As the objects move through a sequence, the state vectors are propagated accordingly by estimating a dense correspondence between consecutive frames. The propagated state vectors are then fused with the target appearance model in order to predict the final target confidence values used for localization. Lastly, the outputs of the predictor and the target model are used to update the state vectors using a convolutional gated recurrent unit (ConvGRU)~\cite{ConvGRU}.

\subsection{Tracking with Scene Propagation}
\label{sec:overview}
Our tracker predictions are based on two cues: (i) appearance in the current frame and (ii) scene information propagated over time.
The appearance model $\tau$ aims to distinguish the target object from the background. By taking the deep feature map $x_t \in \reals^{W \times H \times D}$ extracted from frame $t$ as input, the appearance model $\tau$ predicts a score map $s_t = \tau(x_t) \in \reals^{W \times H}$. Here, the score $s_t(\vecn{r})$ at every spatial location $\vecn{r} \in \Omega \defeq \{0,\dots,W - 1\} \times \{0,\dots,H - 1\}$ denotes the likelihood of that location being the target center. 

The target model has the ability to recover from occlusions and provides long-term robustness. However, it is oblivious to the contents of the surrounding scene. In order to extract such information, our tracker maintains a state vector for every region in the target neighborhood. Concretely, for every spatial location $\vecn{r} \in \Omega$ in the deep feature representation $x_t$, we maintain a $S$-dimensional state vector $h^\vecn{r}$ for that cell location such that $h \in \reals^{W \times H \times S}$. The state vectors contain information about the cell which is beneficial for single target tracking. For example, it can encode whether a particular cell corresponds to the target, background, or is in fact a distractor that looks similar to the target. Note that we do not explicitly enforce any such encoding, but let $h$ be a generic representation whose encoding is trained end-to-end by minimizing a tracking loss.

The state vectors are initialized in the first frame using a small network $\init$ which takes the first frame target annotation $B_0$ as input. The network generates a single-channel label map specifying the target location. This is passed through two convolutional layers to obtain the initial state vectors $h_0 = \init(b_0)$. The state vectors contain localized information specific to their corresponding image regions. Thus, as the objects move through a sequence, we propagate their state vectors accordingly. Given a new frame $t$, we transform the states $h_{t-1}$ from the previous frame locations to the current frame locations. This is performed by our state propagation module $\prop$,
\begin{equation}
\label{eq:state_propagation}
	(\hat{h}_{t-1}, \propconf_t) = \prop(\motionfeat_t, \motionfeat_{t-1}, h_{t-1})
\end{equation}
Here, $\motionfeat_t \in \reals^{W \times H \times D}$ and $\motionfeat_{t-1} \in \reals^{W \times H \times D}$ are the deep feature representations from the current and previous frames, respectively. The output $\hat{h}_{t-1}$ represents the spatially propagated state, compensating for the motions of objects and background in the scene. The propagation reliability map $\propconf_t \in \reals^{W \times H}$ indicates the reliability of the state propagation. That is, a high $\propconf_t(\vecn{r})$ indicates that the state $\hat{h}_{t-1}^\vecn{r}$ at $\vecn{r}$ has been confidently propagated. The reliability map $\propconf_t$ can thus be used to determine whether to trust a propagated state vector $\hat{h}_{t-1}^{\vecn{r}}$ when localizing the target.

\newcommand{\assign}{\leftarrow}
\newcommand{\algcomment}[2]{\hspace{#2mm}{\footnotesize\# #1}}
\begin{algorithm}[b]
	\caption{Tracking loop}
	\begin{algorithmic}[1]
		\Require Image features $\{x_t\}_{t=0}^N$, initial annotation $b_0$, appearance model $\tau$
		\State $h_0 \assign \init(b_0)$ \algcomment{Initialize states}{38}
		\For{$i = 1, \ldots, N$}       \algcomment{For every frame}{27}
		\State $s_t \assign \tau(x_t)$ \algcomment{Apply appearance model}{24}
		\State $(\hat{h}_{t-1}, \propconf_t) \assign \prop(\motionfeat_t, \motionfeat_{t-1}, h_{t-1})$ \algcomment{Propagate states}{5}
		\State $\fusedscore_t \assign P(\hat{h}_{t-1}, \propconf_t, s_t)$ \algcomment{Predict target confidence scores}{4}
		\State $h_t \assign \stateupdate(\hat{h}_{t-1}, \fusedscore_t, s_t)$ \algcomment{Update states}{24}
		\EndFor
	\end{algorithmic}
	\label{alg:tracking_loop}
\end{algorithm}

In order to predict the location of the target object, we utilize \emph{both} the appearance model output $s_t$ and the propagated states $\hat{h}_{t-1}$. The latter captures valuable information about all objects in the scene, which complements the target-centric information contained in the appearance model. We input the propagated state vectors $\hat{h}_{t-1}$, along with the reliability scores $\propconf_t$ and the appearance model prediction $s_t$ to the predictor module $\predictor$. The predictor combines these information to provide the fused target confidence scores $\fusedscore_t$,
\begin{equation}
	\fusedscore_t = P(\hat{h}_{t-1}, \propconf_t, s_t)
\end{equation}
The target is then localized in frame $t$ by selecting the location $\vecn{r^*}$ with the highest score: $\vecn{r^*} = \argmax_{\vecn{r} \in \Omega} \fusedscore_t$. Finally, we use the fused confidence scores $\fusedscore_t$ along with the appearance model output $s_t$ to update the state vectors,
\begin{equation}
\label{eq:state_update}
h_t = \stateupdate(\hat{h}_{t-1}, \fusedscore_t, s_t)
\end{equation}
The recurrent state update module $\stateupdate$ can use the current frame information from the score maps to \eg\ reset an incorrect state vector $\hat{h}_{t-1}^\vecn{r}$, or flag a newly entered object as a distractor. These updated state vectors $h_t$ are then used to track the object in the next frame. Our tracking procedure is detailed in Alg.~\ref{alg:tracking_loop}.

\subsection{State propagation}
\label{sec:state_prop}
The state vectors contain localized information for every region in the target neighborhood. As these regions move through a sequence due to \eg\ object or camera motion, we need to propagate their states accordingly, in order to compensate for their motions. This is done by our state propagation module $\prop$.
The inputs to this module are the deep feature maps $\motionfeat_{t-1}$ and $\motionfeat_t$ extracted from the previous and current frames, respectively. Note that the deep features $\motionfeat$  are not required to be the same as the ones as used for the target model. However, we assume that both feature maps have the same spatial resolution $W \times H$.

In order to propagate the states from the previous frame to the current frame locations, we first compute a dense correspondence between the two frames. We represent this correspondence as a probability distribution $\corr$, where $\corrat{r'}{r}$ is the probability that location $\vecn{r} \in \Omega$ in the current frame originated from $\vecn{r'} \in \Omega$ in the previous frame. The dense correspondence is estimated by constructing a 4D cost volume $\cv \in \reals^{W \times H \times W \times H}$, as is commonly done in optical flow approaches~\cite{Flownet,PWCNet,DCFlow}. The cost volume contains a matching cost between every image location pair from the previous and current frame. The element $\cvat{r'}{r}$ in the cost volume is obtained by computing the correlation between $3 \times 3$ windows centered at $\vecn{r'}$ in the previous frame features $\motionfeat_{t-1}$ and $\vecn{r}$ in the current frame features $\motionfeat_t$. For computational efficiency, we only construct a partial cost volume by assuming a maximal displacement of $d_\text{max}$ for every feature cell.

We process the cost volume through a network module to obtain robust dense correspondences. We pass the cost volume slice $\cv_{\vecn{r'}}(\vecn{r}) \in \reals^{W \times H}$ for every cell $\vecn{r'}$ in the previous frame, through two convolutional blocks in order to obtain processed matching costs $\phi(\vecn{r'}, \vecn{r})$. Next, we take the softmax of this output over the current frame locations to get an initial correspondence $\phi'(\vecn{r'}, \vecn{r}) = \frac{\expo{\phi(\vecn{r'}, \vecn{r})}}{\sum_{\vecn{r''} \in \Omega}\expo{\phi(\vecn{r'}, \vecn{r''})}}$. The softmax operation aggregates information over the current frame dimension and provides a soft association of locations between the two frames. In order to also integrate information over the previous frame locations, we pass $\phi'$ through two more convolutional blocks and take softmax over the previous frame locations. This provides the required probability distribution $\corrat{r'}{r}$ at each current frame location $\vecn{r}$.

The estimated correspondence $\corrat{r'}{r}$ between the frames can now be used to determine the propagated state vector $\hat{h}_{t-1}^{\vecn{r}}$ at a current frame location $\vecn{r}$ by evaluating the following expectation over the previous frame state vectors.
\begin{equation}
\label{eq:state_prop}
\hat{h}_{t-1}^{\vecn{r}} = \sum_{\vecn{r'} \in \Omega} h_{t-1}^{\vecn{r'}} \corrat{\vecn{r'}}{\vecn{r}} \,.
\end{equation} 
When using the propagated state vectors $\hat{h}_{t-1}$ for target localization, it is also helpful to know if a particular state vector is valid \ie\ if it has been correctly propagated from the previous frame. We can estimate this reliability $\propconf_{t}^{\vecn{r}}$ at each location $\vecn{r}$ using the correspondence probability distribution $\corrat{r'}{r}$ for that location. A single mode in $\corrat{r'}{r}$ indicates that we are confident about the source of the location $\vecn{r}$ in the previous frame. A uniformly distributed $\corrat{r'}{r}$ on the other hand implies uncertainty. In such a scenario, the expectation \ref{eq:state_prop} reduces to a simple average over the previous frame state vectors $h_{t-1}^{\vecn{r'}}$, leading to an unreliable $\hat{h}_{t-1}^{\vecn{r}}$. Thus, we use the negation of the shannon entropy of the distribution $\corrat{r'}{r}$ to obtain the reliability score $\propconf_{t}^{\vecn{r}}$ for state $\hat{h}_{t-1}^{\vecn{r}}$, 
\begin{equation}
\label{eq:prop_confidence}
	\propconf_{t}^{\vecn{r}} = \sum_{\vecn{r'} \in \Omega} \corrat{r'}{r}\log(\corrat{r'}{r})
\end{equation}
The reliability $\propconf_{t}^{\vecn{r}}$ is then be used to determine whether to trust the state $\hat{h}_{t-1}^{\vecn{r}}$ when predicting the final target confidence scores. 
 
\subsection{Target Confidence Score Prediction}
\label{sec:prediction}
In this section, we describe our predictor module $P$ which determines the target location in the current frame. We utilize both the appearance model output $s_t$ and the scene information encoded by $\hat{h}_{t-1}$ in order to localize the target. The appearance model score $s_t^{\vecn{r}}$ indicates whether a location $\vecn{r}$ is target or background, based on the appearance in the current frame only. The state vector $\hat{h}_{t-1}^{\vecn{r}}$ on the other hand contains past information for \emph{every} location $\vecn{r}$. It can, for instance, encode whether the cell $\vecn{r}$ was classified as target or background in the previous frame, how certain was the tracker prediction for that location, and so on. The corresponding reliability score $\propconf_{t}^{\vecn{r}}$ further indicates if the state vector $\hat{h}_{t-1}^{\vecn{r}}$ is reliable or not. This can be used to determine how much weight to give to the state vector information when determining the target location.

The predictor module $\predictor$ is trained to effectively combine the information from $s_t$, $\hat{h}_{t-1}$, and $\propconf_{t}$ to output the final target confidence score $\fusedscore_t \in \reals^{W \times H}$. We concatenate the appearance model output $s_t$, the propagated state vectors $\hat{h}_{t-1}$, and the state reliability scores $\propconf_t$ along the channel dimension, and pass the resulting tensor through two convolutional blocks. The output is then mapped to the range $[0, 1]$ by passing it through a sigmoid layer to obtain the intermediate scores $\hat{\fusedscore}_t$. While it is possible to use this score directly, it is not reliable in case occlusions. This is because the state vectors corresponding to the target can leak into the occluding object, especially when two objects cross each other slowly. The fused scores can be thus be corrupted in such cases. In order to handle this, we pass $\hat{\fusedscore}_t$ through another layer which masks the regions from the score map $\hat{\fusedscore}_t$ where the appearance model score $s_t$ is less than a threshold $\mu$. Thus, we let the appearance model override the predictor output in case of occlusions. The final score map $\fusedscore_t$ is thus obtained as $\fusedscore_t = \hat{\fusedscore_t} \cdot \mathbbm{1}_{s_t > \mu}$. Here, $\mathbbm{1}_{s_t > \mu}$ is an indicator function which evaluates to $1$ when $s_t > \mu$ and is $0$ otherwise and $\cdot$ denotes elementwise product. Note that the masking operation is differentiable and is implemented inside the network.

\subsection{State update}
\label{sec:state_update}
While the state propagation described in Section~\ref{sec:state_prop} maps the state to the new frame, it does not update it with new information about the scene. This is accomplished by a recurrent neural network module, which evolves the state in each time step.
As tracking information about the scene, we input the scores $s_t$ and $\fusedscore_t$ obtained from the appearance model $\tau$ and the predictor module $P$, respectively. The update module can thus \eg\ mark a new distractor object which entered the scene or correct corrupted states which have been incorrectly propagated.
This state update is performed by the recurrent module $\stateupdate$ (eq.\ \ref{eq:state_update}). 

\begin{figure*}[t]
	\centering%
	\newcommand{\wid}{0.95\textwidth}%
	\includegraphics*[trim = 0 0 0 0, width = \wid]{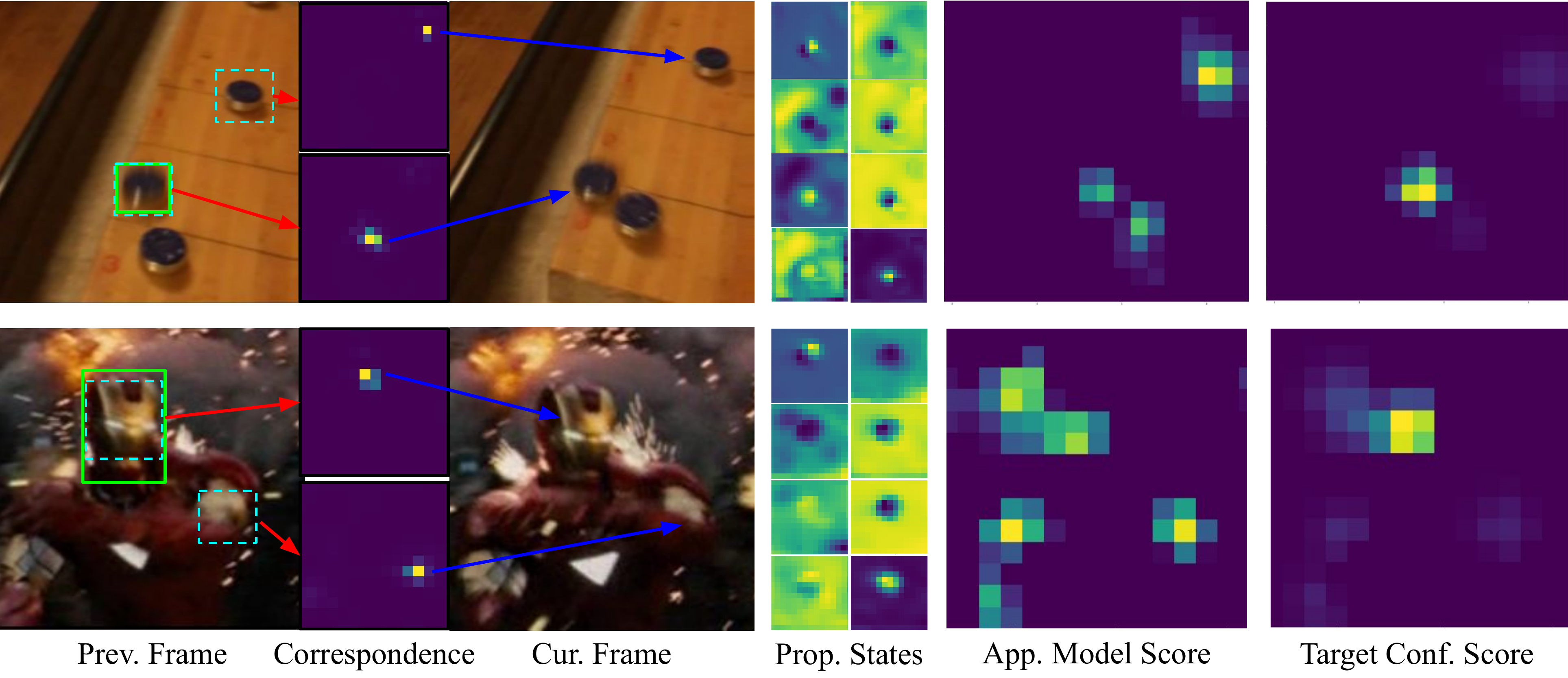}\vspace{-4mm}%
	\caption{Visualization of intermediate representations used for tracking on two example sequences. The green box in the previous frame (first column) denotes the target to be tracked. For every location in the current frame (third column), we plot the estimated correspondence with the marked region in the previous frame (second column). The states propagated to the current frame using the estimated correspondence are plotted channel-wise in the fourth column. The appearance model score (fifth column) fails to correctly localize the target in both cases due to the presence of distractors. In contrast, our approach can correctly handle these challenging scenarios and provides robust target confidence scores (last column) by exploiting the propagated scene information.}%
	\label{fig:qual_fig}%
	\vspace{-4mm}
\end{figure*}

The update module $\stateupdate$ contains a convolutional gated recurrent unit (ConvGRU)~\cite{ConvGRU,GRU}. We concatenate the scores $\fusedscore_t$ and $s_t$ along with their maximum values in order to obtain the input $f_t \in \reals^{W \times H \times 4}$ to the ConvGRU. The propagated states from the previous frame $\hat{h}_{t-1}$ are treated as the hidden states of the ConvGRU from the previous time step. The ConvGRU then updates the previous states using the current frame observation $f_t$ to provide the new states $h_t$. A visualization of the representations used by our tracker is shown in Fig.~\ref{fig:qual_fig}.

\subsection{Target Appearance Model}
Our approach can be integrated with any tracking appearance model. In this work, we use the DiMP tracker~\cite{DiMP} as our target model component, due to its strong performance. DiMP is an end-to-end trainable tracking architecture that predicts the appearance model $\tau_w$, parametrized by the weights $w$ of a single convolutional layer. The network integrates an optimization module that minimizes the following discriminative learning loss,
\begin{equation}
\label{eq:dimp_loss}
L(w) = \frac{1}{|S_\text{train}|} \sum_{(x,c) \in S_\text{train}} \|r(\tau_w(x), c)\|^2 + \|\lambda w\|^2 \,. 
\end{equation} 
Here, $\lambda$ is the regularization parameter. The training set $S_\text{train}=\{(x_j, c_j)\}_{j=1}^n$ consists of deep feature maps $x_j$ extracted from the training images, and the corresponding target annotations $c_j$. The residual function $r(s, c)$ computes the error between the tracker prediction $s = \tau_w(x)$ and the groundtruth. The training set is constructed in the initial frame by employing different data augmentation strategies. We refer to \cite{DiMP} for more details about the DiMP tracker.

\subsection{Offline Training}
\label{sec:training}
In order to train our architecture, it is important to simulate the tracking scenario. This is needed to ensure that the network can learn to effectively propagate the scene information over time and determine how to best fuse it with the appearance model output. Thus, we train our network using video sequences. We first sample a set of $N_\text{train}$ frames from a video, which we use to construct the appearance model $\tau$. We then sample a sub-sequence $V = \{(I_t, b_t)\}_{t=0}^{N_\text{seq} - 1}$ consisting of $N_\text{seq}$ consecutive frames $I_t$ along with their corresponding target annotation $b_t$. We apply our network on this sequence data, as it would be during tracking. We first obtain the initial state $h_0 = \init(b_0)$ using the state initializer $\init$. The states are then propagated to the next frame (Sec. \ref{sec:state_prop}), used to predict the target scores $\fusedscore_t$ (Sec. \ref{sec:prediction}), and finally updated using the predicted scores (Sec. \ref{sec:state_update}). This procedure is repeated until the end of the sequence and the training loss is computed by evaluating the tracker performance over the whole sequence. 

In order to obtain the tracking loss $L$, we first compute the prediction error $L_t^\text{pred}$ for every frame $t$ using the standard least-squares loss,
\begin{equation}
\label{eq:pred_loss}
L_t^\text{pred} = \left\|\fusedscore_t - z_t\right\|^2
\end{equation}
Here, $z_t$ is a label function, which we set to a Gaussian centered at the target. We also compute a prediction error $L_t^\text{pred, raw}$ using the raw score map $\hat{\fusedscore}_t$ predicted by $\predictor$ in order to obtain extra training supervision. To aid the learning of the state vectors and the propagation module $\prop$, we add an additional auxiliary task. We use a small network head to predict whether a state vector $h_{t-1}^{\vecn{r}}$ corresponds to the target or background. This prediction is penalized using a binary cross entry loss to obtain $L_t^\text{state}$. The network head is also applied on the propagated state vectors $\hat{h}_{t-1}^{\vecn{r}}$ to get $L_t^\text{state, prop}$. This loss provides a direct supervisory signal to the propagation module $\prop$.

Our final tracking loss $L$ is obtained as the weighted sum of the above individual losses over the whole sequence,
\begin{equation}
\label{eq:total_loss}
L = \frac{1}{N_\text{seq} - 1} \!\!\!\sum_{t=1}^{N_\text{seq} - 1} \!L_t^\text{pred} + \alpha L_t^\text{pred, raw} + \beta (L_t^\text{state} + L_t^\text{state, prop}) \,.
\end{equation}
The hyper-parameters $\alpha$ and $\beta$ determine the impact of the different losses. Note that the scores $s_t$ predicted by the appearance model can itself localize the target correctly in a majority of the cases. Thus, there is a risk that the predictor module only learns to rely on the target model scores $s_t$. To avoid this, we randomly add distractor peaks to the scores $s_t$ during training to encourage the predictor to utilize the scene information encoded by the state vectors. 

\subsection{Implementation details}
\label{sec:details}
We use a pre-trained DiMP model with ResNet-50~\cite{Resnet} backbone from \cite{pytracking} as our target appearance model. We use the block 4 features from the same backbone network as input to the state propagation module $\prop$. For computational efficiency, our tracker does not process the full input image. Instead, we crop a square region containing the target, with an area $5^2$ times that of the target. The cropped search region is resized to $288 \times 288$ size, and passed to the network. We use $S = 8$ dimensional state vectors to encode the scene information. The threshold $\mu$ in the predictor $\predictor$ is set to $0.05$.

We use the training splits of TrackingNet~\cite{TrackingNet}, LaSOT~\cite{LaSOT}, and GOT-10k~\cite{GOT10k} datasets to train our network.
Within a sequence, we perturb the target position and scale in every frame in order to avoid learning any motion bias. While our network is end-to-end trainable, we do not fine-tune the weights for the backbone network due to GPU memory constraints. 
Our network is trained for $40$ epochs, with $1500$ sub-sequences in each epoch. We use the ADAM~\cite{ADAM} optimizer with an initial learning rate of $10^{-2}$, which is reduced by a factor of $5$ every $20$ epochs. We use $N_\text{train} = 3$ frames to construct the appearance model while the sub-sequence length is set to $N_\text{seq} = 50$. The loss weights are set to $\alpha = \beta = 0.1$.

During online tracking, we use a simple heuristic to determine target loss.  In case the fused confidence score $\fusedscore_t$ peak is smaller than a threshold ($0.05$), we infer that the target is lost and do not update the state vectors in this case. We impose a prior on the target motion by applying a window function on the appearance model prediction $s_t$ input to $\predictor$, as well as the output target confidence score $\fusedscore_t$. We also handle any possible drift in the target confidence scores. In case the appearance model scores $s_t$ and target confidence score $\fusedscore_t$ only have small offset in their peaks, we use the appearance model score to determine the target location as it is more resistant to drift. After determining the target location, we use the bounding box estimation branch in DiMP to obtain the target box. 

\section{Experiments}
We evaluate our proposed tracking architecture on five tracking benchmarks: VOT2018~\cite{VOT2018}, GOT-10k~\cite{GOT10k}, TrackingNet~\cite{TrackingNet}, OTB-100~\cite{OTB2015}, and NFS~\cite{NfS}. Detailed results are provided in the supplementary material. Our tracker operates at around $20$ FPS on a single Nvidia RTX 2080 GPU. The complete training and inference code will be released upon publication.

\subsection{Ablation study}
We conduct an ablation study to analyze the impact of each component in our tracking architecture. We perform experiments on the combined NFS~\cite{NfS} and OTB-100~\cite{OTB2015} datasets consisting of $200$ challenging videos. The trackers are evaluated using the overlap precision (OP) metric. The overlap precision OP$_T$ denotes the percentage of frames where the intersection-over-union (IoU) overlap between the tracker prediction and the groundtruth box is higher than a threshold $T$. The OP scores over a range of thresholds $[0, 1]$ are averaged to obtain the area-under-the-curve (AUC) score. We report the AUC and OP$_{0.5}$ scores for each tracker. Due to the stochastic nature of our appearance model, all results are reported as the average over $5$ runs. Unless stated otherwise, we use the same training procedure and settings mentioned in sections \ref{sec:training} and \ref{sec:details}, respectively, to train all trackers evaluated in this section.

\begin{table}[!t]
	\centering\vspace{0mm}
	\caption{Impact of each component in our tracking architecture on the combined NFS and OTB-100 datasets. Compared to using only the appearance model, our approach integrating scene knowledge, provides a significant $1.3\%$ improvement in AUC score.}\vspace{-2mm}
	\resizebox{\columnwidth}{!}{%
		\begin{tabular}{l@{~}c@{~~}c@{~~}c@{~~}c@{~~}c@{~~}}
\toprule
&\textbf{Ours}&Only Appearance&No State&No Propagation&No Appearance\\
&&Model $\tau$&Propagation $\prop$&Reliability $\propconf_{t}$&Model $\tau$\\\midrule
AUC(\%)&\textbf{66.4}&65.1&64.9&66.1&49.2\\
OP$_{0.5}$&\textbf{83.5}&81.9&81.2&82.9&60.1\\\bottomrule

\end{tabular}

	}%
	\label{tab:ablation}%
	\vspace{-4mm}
\end{table}

\parsection{Impact of scene information} In order to study the impact of integrating scene information for tracking, we compare our approach with a tracker only employing  target appearance model $\tau$. This version is equivalent to the standard DiMP-50~\cite{DiMP}. The results are reported in Tab. \ref{tab:ablation}. Note that our appearance model is itself a state-of-the-art tracker, obtaining the best tracking performance on multiple tracking benchmarks~\cite{DiMP}. Compared to using only the appearance model, our approach exploiting scene information provides an improvement of $1.3\%$ and $1.6\%$ in AUC and OP$_{0.5}$ scores, respectively. These results clearly demonstrates that scene knowledge contains complementary information that benefits tracking performance, even when integrated with a strong appearance model.

\parsection{Impact of state propagation} Here, we analyze the impact of state propagation module (Sec.~\ref{sec:state_prop}), which maps the localized states between frames by generating dense correspondences. This is performed by replacing the propagation module $\prop$ in \eqref{eq:state_propagation} and \eqref{eq:state_prop} with an identity mapping $\hat{h}_{t-1}=h_{t-1}$. That is, the states are no longer explicitly tracked by computing correspondences between frames.
The results for this experiment are shown in Table~\ref{tab:ablation}. 
Interestingly, the approach without state propagation performs slightly worse ($0.2\%$ in AUC) than the network using only the appearance model. This shows that state propagation between frames is critical in order to exploit the localized scene information.

\parsection{Impact of propagation reliability} Here, we study the impact of the propagation reliability score $\propconf_{t}$ for confidence score prediction. We compare our approach with a baseline tracker which does not utilize $\propconf_{t}$. The results indicate that using reliability score $\propconf_{t}$ is beneficial, leading to a $+0.3\%$ AUC improvement.

\parsection{Impact of appearance model} Our architecture utilizes the propagated scene information to \emph{complement} the frame-by-frame prediction performed by the target appearance model. By design, our tracker relies on the appearance model to provide long-term robustness in case of \eg\ occlusions, and thus is not suited to be used without it. However, for completeness, we evaluate a version of our tracker which does not utilize any appearance model. That is, we only use the propagated states $\hat{h}_{t-1}$, and the reliability score $\propconf_{t}$ in order to track the target. As expected, not using an appearance model substantially deteriorates the performance by over $17\%$ in AUC score. 

\subsection{State-of-the-art Comparison}

\begin{figure*}[t]
	\newcommand{\wid}{0.33\textwidth}%
	\centering\vspace{-5mm}%
	\subfloat[GOT-10k\label{fig:sota_got}]{\includegraphics[width = \wid]{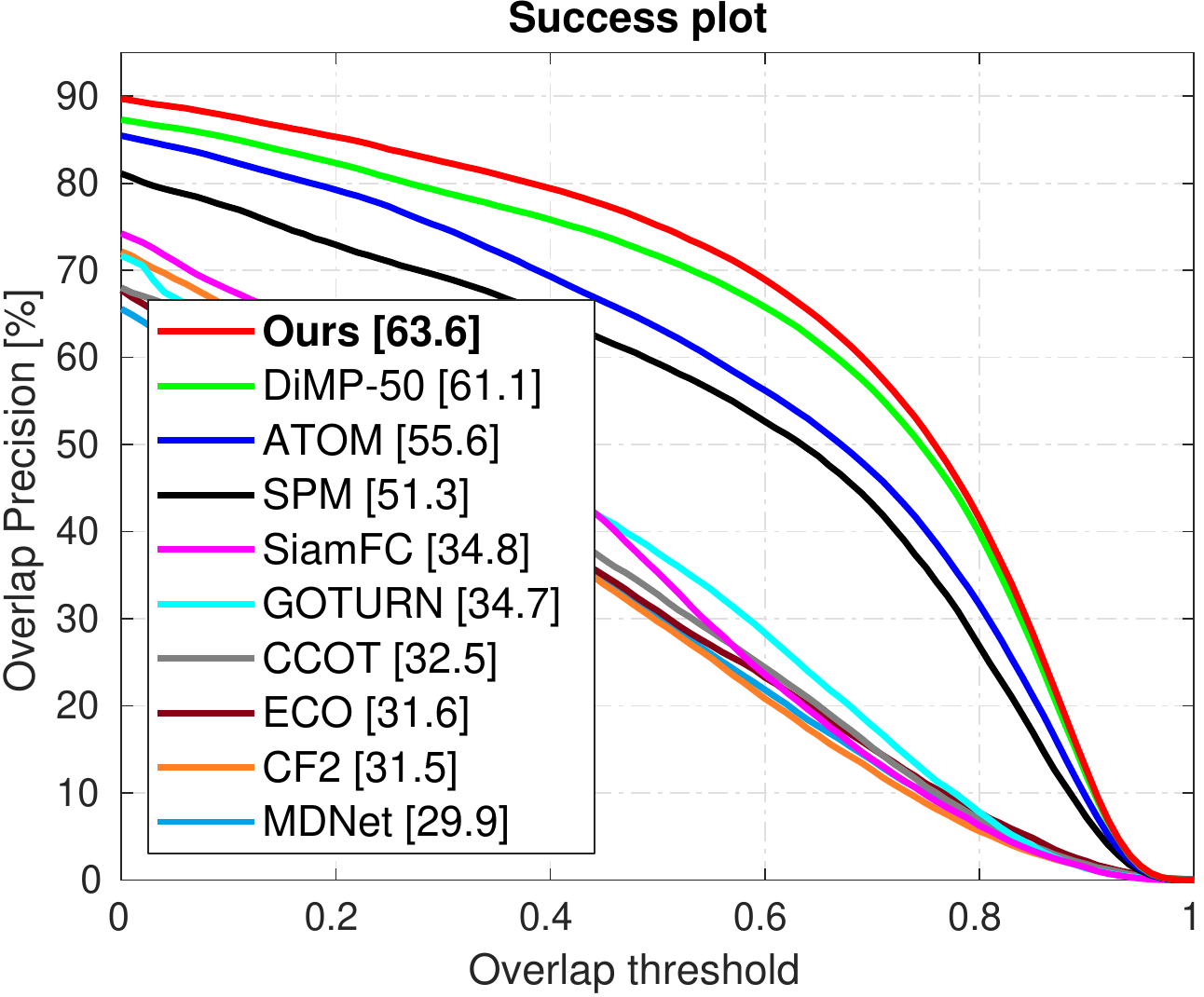}}%
	\subfloat[OTB-100\label{fig:sota_otb}]{\includegraphics[width = \wid]{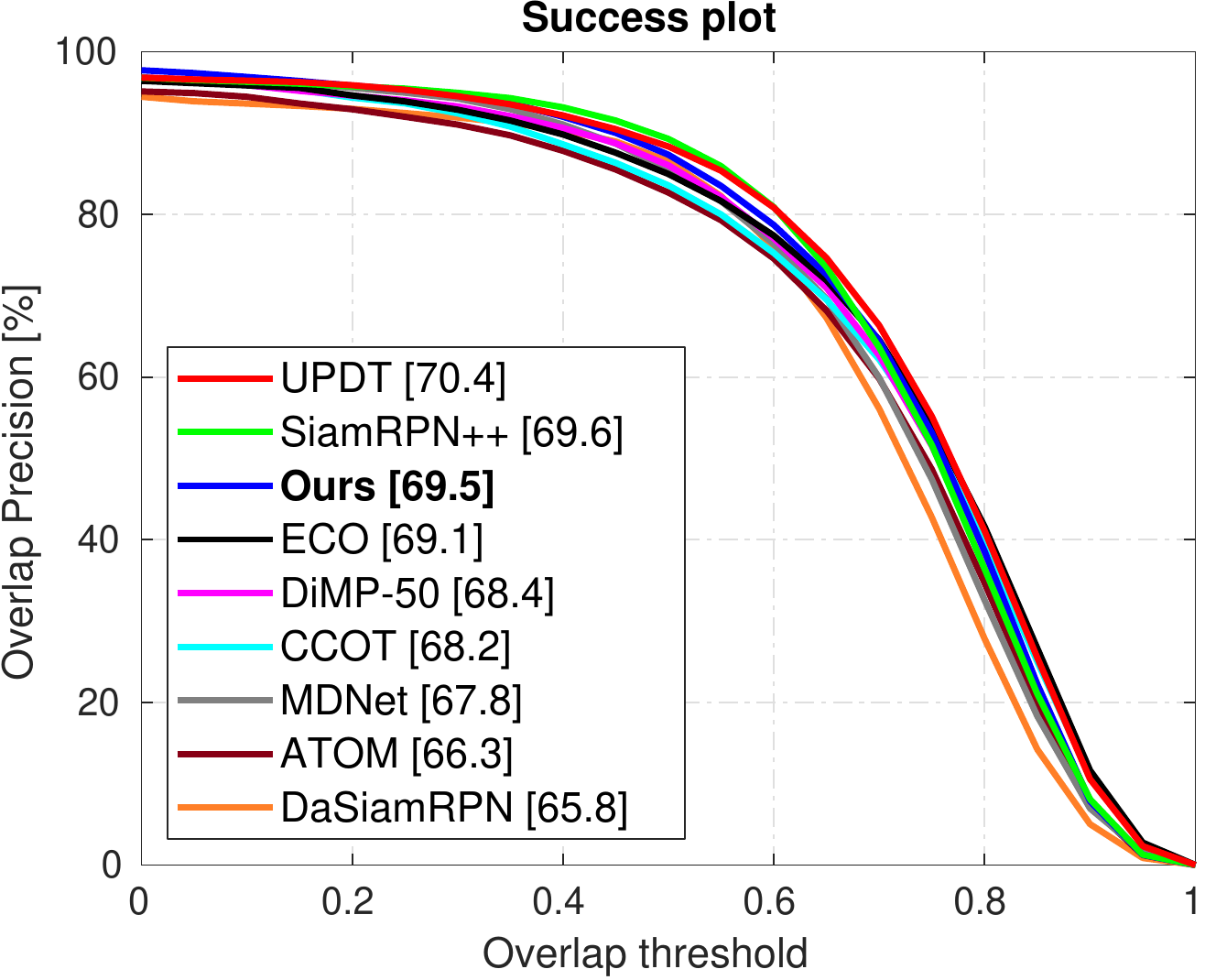}}%
	\subfloat[NFS\label{fig:sota_nfs}]{\includegraphics[width = \wid]{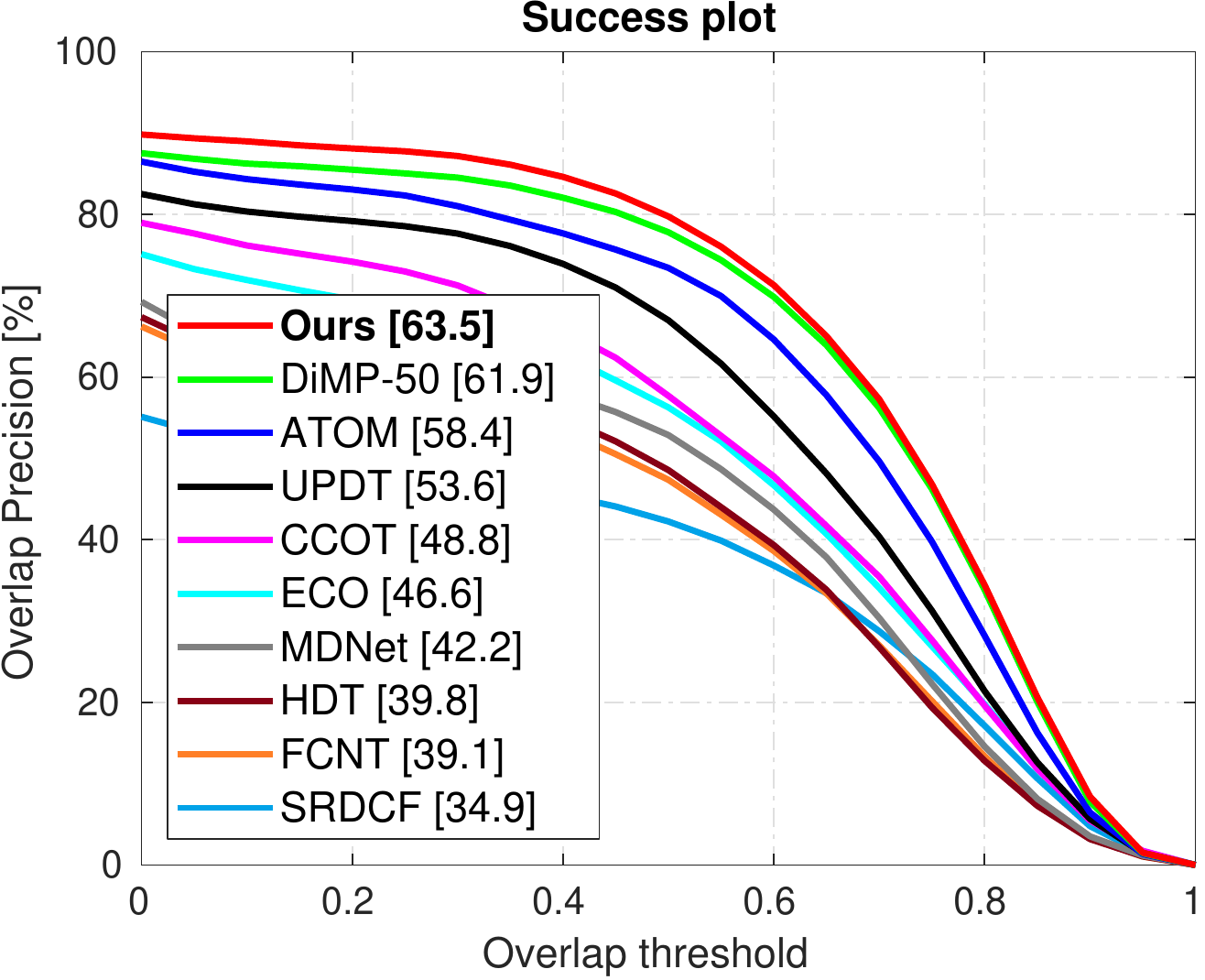}}%
	\caption{Success plots on GOT-10k (a), OTB-100 (b) and NFS (c). 
		The AUC scores are shown in legend. 
		Our approach obtains the best results on both GOT-10k and NFS datasets, outperforming the previous best method by $2.5\%$ and $1.6\%$ AUC, respectively. 
	}%
	\label{fig:sotanfsotb}\vspace{-3mm}
\end{figure*}
In this section, we compare our proposed tracker with the state-of-the-art approaches on five tracking benchmarks.

\begin{table}[!b]
	\centering\vspace{-4mm}
	\caption{State-of-the-art comparison on the VOT2018 in terms of expected average overlap (EAO), accuracy and robustness. Our approach obtains the best EAO score, outperforming the previous best approach DiMP-50 with a EAO relative gain of $5.0\%$.
	}\vspace{-2mm}
	\resizebox{\textwidth}{!}{%
		\begin{tabular}{l@{~}c@{~~}c@{~~}c@{~~}c@{~~}c@{~~}c@{~~}c@{~~}c@{~~}c@{~~}c@{~~}}
	\toprule
	&DRT&RCO&UPDT&DaSiam-&MFT&LADCF&ATOM&SiamRPN++&DiMP-50&\textbf{Ours}\\
	&\cite{DRT}&\cite{VOT2018}&\cite{BhatECCV2018}&RPN \cite{DaSiamRPN}&\cite{VOT2018}&\cite{LADCF}&\cite{ATOM}&\cite{SiamRPN++}&\cite{DiMP}&\\\midrule
	EAO&0.356&0.376&0.378&0.383&0.385&0.389&0.401&0.414&\textbf{\textcolor{blue}{0.440}}&\textbf{\textcolor{red}{0.462}}\\
	Robustness&0.201&0.155&0.184&0.276&\textbf{\textcolor{red}{0.140}}&0.159&0.204&0.234&0.153&\textbf{\textcolor{blue}{0.143}}\\
	Accuracy&0.519&0.507&0.536&0.586&0.505&0.503&0.590&\textbf{\textcolor{blue}{0.600}}&0.597&\textbf{\textcolor{red}{0.609}}\\\bottomrule
\end{tabular}

	}\vspace{1mm}%
	\label{tab:vot}%
	\vspace{-1mm}
\end{table}
\parsection{VOT2018~\cite{VOT2018}} We evaluate our approach on the VOT2018 dataset consisting of $60$ videos. 
The trackers are compared using the measures robustness and accuracy. Robustness indicates the number of tracking failures, while accuracy denotes the average overlap between tracker prediction and the ground-truth box. Both these measures are combined into a single expected average overlap (EAO) score.
We compare our proposed tracker with the state-of-the-art approaches. Results are shown in Tab.~\ref{tab:vot}. Note that all top ranked approaches on VOT2018 only utilize a target appearance model for tracking. In contrast, our approach also exploits explicit knowledge about other objects in the scene. In terms of the overall EAO score, our approach outperforms the previous best method DiMP-50 with a large margin, achieving a relative gain of $5.0\%$ in EAO.

\parsection{GOT10k~\cite{GOT10k}} 
\begin{table}[t]
    \caption{State-of-the-art comparison on the GOT-10k test set in terms of average overlap (AO) and success rates (SR) at overlap thresholds $0.5$ and $0.75$. Our approach obtains the best results in all three measures, achieving an AO score of $63.6$.
	}%
	\centering\vspace{-2mm}
	\resizebox{\columnwidth}{!}{%
		\begin{tabular}{l@{~}c@{~~}c@{~~}c@{~~}c@{~~}c@{~~}c@{~~}c@{~~}c@{~~}c@{~~}c@{~~}}
	\toprule
	&MDNet&CF2&ECO&CCOT&GOTURN&SiamFC&SiamFCv2&ATOM&DiMP-50&\textbf{Ours}\\
	&\cite{MDNet}&\cite{HCF_ICCV15}&\cite{DanelljanCVPR2017}&\cite{DanelljanECCV2016}&\cite{Held2016gotrun}&\cite{SiameseFC}&\cite{Valmadre2017cvpr}&\cite{ATOM}&\cite{DiMP}&\\\midrule
	SR$_{0.50}$ (\%)&30.3&29.7&30.9&32.8&37.5&35.3&40.4&63.4&\textbf{\textcolor{blue}{71.7}}&\textbf{\textcolor{red}{75.1}}\\
	SR$_{0.75}$ (\%)&9.9&8.8&11.1&10.7&12.4&9.8&14.4&40.2&\textbf{\textcolor{blue}{49.2}}&\textbf{\textcolor{red}{51.5}}\\
	AO (\%)&29.9&31.5&31.6&32.5&34.7&34.8&37.4&55.6&\textbf{\textcolor{blue}{61.1}}&\textbf{\textcolor{red}{63.6}}\\\bottomrule
\end{tabular}

	}\vspace{1mm}%
	\label{tab:got_sota}%
	\vspace{-2mm}
\end{table}
This is a recently introduced large scale dataset consisting of over $10,000$ videos. In contrast to other datasets, trackers are restricted to use only the train split of the dataset in order to train networks \ie\ use of external training data is forbidden. Accordingly, we train our network using only the train split. We ensure that our appearance model $\tau$ is also trained using only the train split. The results are reported on the test split consisting of $180$ videos. 
The results, in terms of average overlap (AO) and success rates at overlap thresholds $0.5$ and $0.75$ are shown in Table~\ref{tab:got_sota}, while Figure~\ref{fig:sota_got} shows the success plots. Among the previous methods, the appearance model used by our tracker, namely  DiMP-50, obtains the best results. Our approach, integrating scene information for tracking, significantly outperforms DiMP-50, setting a new state-of-the-art on this datatset. Our tracker achieves an AO score of $63.6$, a relative improvement of $4.1\%$ over the previous best method. These results clearly show the benefits of exploiting scene knowledge for tracking.

\parsection{TrackingNet~\cite{TrackingNet}} The large scale TrackingNet dataset consists of over $30,000$ videos sampled from YouTube. We report results on the test split, consisting of $511$ videos. The results in terms of precision, normalized precision, and success are shown in Table~\ref{tab:trackingnet_sota}. The baseline approach DiMP-50 already achieves the best results with an AUC of $74.0$. Our approach achieves a similar performance to the baseline, showing that it generalizes well to such real world videos.
 
\begin{table}[t]
	\caption{State-of-the-art comparison on the TrackingNet test set in terms of precision, normalized precision, and success. Our approach performs similarly to previous best method DiMP-50, achieving an AUC score of $74.0\%$.}%
	\centering\vspace{-2mm}
	\resizebox{\columnwidth}{!}{%
		\begin{tabular}{l@{~}c@{~~}c@{~~}c@{~~}c@{~~}c@{~~}c@{~~}c@{~~}c@{~~}c@{~~}c@{~~}}
	\toprule
	&ECO&SiamFC&CFNet&MDNet&UPDT&DaSiam-&ATOM&SiamRPN++&DiMP-50&\textbf{Ours}\\
	&\cite{DanelljanCVPR2017}&\cite{SiameseFC}&\cite{Valmadre2017cvpr}&\cite{MDNet}&\cite{BhatECCV2018}&RPN \cite{DaSiamRPN}&\cite{ATOM}&\cite{SiamRPN++}&\cite{DiMP}&\\\midrule
	Precision (\%)&49.2&53.3&53.3&56.5&55.7&59.1&64.8&\textbf{\textcolor{red}{69.4}}&68.7&\textbf{\textcolor{blue}{68.8}}\\
	Norm.\ Prec.\ (\%)&61.8&66.6&65.4&70.5&70.2&73.3&77.1&\textbf{\textcolor{blue}{80.0}}&\textbf{\textcolor{red}{80.1}}&\textbf{\textcolor{blue}{80.0}}\\
	Success (AUC) (\%)&55.4&57.1&57.8&60.6&61.1&63.8&70.3&\textbf{\textcolor{blue}{73.3}}&\textbf{\textcolor{red}{74.0}}&\textbf{\textcolor{red}{74.0}}\\\bottomrule
\end{tabular}

	}%
	\label{tab:trackingnet_sota}%
	\vspace{-5mm}
\end{table}

\parsection{OTB-100~\cite{OTB2015}} Figure~\ref{fig:sota_otb} shows the success plots over all the $100$ videos. Discriminative correlation filter based UPDT~\cite{BhatECCV2018} tracker achieves the best results with an AUC score of $70.4$. Our approach obtains results comparable with the state-of-the-art, while outperforming the baseline DiMP-50 by over $1\%$ in AUC.

\parsection{NFS~\cite{NfS}} The need for speed dataset consists of $100$ challenging videos captured using a high frame rate ($240$ FPS) camera. We evaluate our approach on the downsampled $30$ FPS version of this dataset. The success plots over all the $100$ videos are shown in Fig.~\ref{fig:sota_nfs}. Among previous methods, our appearance model DiMP-50 obtains the best results. Our approach significantly outperforms DiMP-50 with a relative gain of $2.6\%$, achieving $63.5\%$ AUC score.
\section{Conclusions}
We propose a novel tracking architecture which can exploit the scene information to improve tracking performance. Our tracker represents the scene information as dense localized state vectors. These state vectors are propagated through the sequence and combined with the appearance model output to localize the target. 
We evaluate the proposed approach on 5 tracking benchmarks. Our tracker sets a new state-of-the-art on 3 of these benchmarks, demonstrating the benefits of exploiting scene information for tracking.

\noindent\textbf{Acknowledgments}: This work was supported by a Huawei Technologies Oy (Finland) project, the ETH Z\"urich Fund (OK), an Amazon AWS grant, and an Nvidia hardware grant.

\bibliographystyle{splncs04}
\bibliography{references}

\clearpage

\begin{center}
	\textbf{\large Supplementary Material}
\end{center}

\newcommand{\charfunc}{\mathbbm{1}}

The supplementary material provides additional details about the network architecture and results. In Section \ref{sec:network}, we provide details about our tracking architecture. Section~\ref{sec:VOT2018} contains detailed results on the VOT2018 dataset, while Section~\ref{sec:qual} provides qualitative comparison of our approach with the baseline tracker DiMP-50 \cite{DiMP}. 

\section{Network details}
\label{sec:network}
In this section, we provide more details about our tracking architecture. 

\parsection{State initializer $\init$}
Given the first frame target annotation $B_0$ as input, the initializer network $\init$ first generates a single-channel label map specifying the target center. We use a Gaussian function to generate this label map. The label map is passed through a single convolutional layer with $3\times3$ kernels. The output is then passed through a tanh activation to obtain the initial state vectors.

\parsection{State propagation}
We use the features from the fourth convolutional block of ResNet-50 \cite{Resnet}, having a spatial stride of 16, to construct our cost volume. Our network can process images of any input resolution. However, in all our experiments, we resize the input search region crop to $288 \times 288$ for convenience. Thus the features $\motionfeat$ used for computing the cost volume have the size $W = H = 18$, with $D_m = 1024$ channels. The maximal displacement $d_\text{max}$ for cost volume computation is set to $9$.

The network architecture used to map the raw cost volume slices to obtain the processed matching costs $\phi$ is shown in Table~\ref{tab:cv_mapping}. Note that the network weights are shared for all cost volume slices. We use an identical network architecture to process the initial correspondence $\phi'$.

\parsection{Target Confidence Score Prediction}
The network architecture for our predictor module $\predictor$ is shown in Table \ref{tab:predictor}.

\parsection{State update} The state update module $\stateupdate$ contain a convolutional gated recurrent unit (ConvGRU) \cite{ConvGRU} which performs the state updates. The input $f_t \in \reals^{W \times H \times 4}$ to the ConvGRU is obtained by concatenating the target confidence scores $\fusedscore_t \in \reals^{W \times H \times 1}$ and the appearance model output $s_t \in \reals^{W \times H \times 1}$, along with their maximum values along the third dimension. The propagated state vectors $\hat{h}_{t-1} \in \reals^{W \times H \times S}$ are treated as the hidden states of the ConvGRU from the previous time-step. We use the standard update equations for ConvGRU,
\begin{subequations}
	\begin{align}
	& z_t = \sigma\big(\text{Conv}(f_t \oplus \hat{h}_{t-1})\big)\\
	& r_t = \sigma\big(\text{Conv}(f_t \oplus \hat{h}_{t-1})\big)\\
	& \tilde{h_t} = \text{tanh}\big(\text{Conv}(f_t \oplus (r_t \odot \hat{h}_{t-1}))\big)\\
	& h_t = (1 - z_t) \odot \hat{h}_{t-1} + z_t \odot \tilde{h_t} \,.
	\end{align}
\end{subequations}
Here, $\oplus$ denotes concatenation of the feature maps along the third dimension, while $\odot$ denotes element-wise product. $\sigma$ and tanh denote the sigmoid and hyperbolic tangent activation functions, respectively. We use $3 \times 3$ kernels for all the convolution layers, represented by Conv.

\begin{table}[!t]
	\centering\vspace{-1mm}
	\resizebox{0.5\columnwidth}{!}{%
		\begin{tabular}{l@{~~}c@{~~}c@{~~}}
\toprule
Layer&Operation&Output size\\\midrule
1&Conv + BN + ReLU&$18\times18\times8$\\
2&Conv + BN&$18\times18\times1$\\\bottomrule

\end{tabular}

	}\vspace{1mm}%
	\caption{The network architecture used to process cost volume slices. The network takes individual cost volume slices (size $18\times18\times1$) as input. All convolutional layers use $3\times3$ kernels. BN denotes batch normalization \cite{BatchNorm}.}
	\label{tab:cv_mapping}%
	\vspace{-1mm}
\end{table}
\begin{table}[!t]
	\centering\vspace{-1mm}
	\resizebox{0.5\columnwidth}{!}{%
		\begin{tabular}{l@{~~}c@{~~}c@{~~}}
\toprule
Layer&Operation&Output size\\\midrule
1&Conv + ReLU&$18\times18\times16$\\
2&Conv + Sigmoid&$18\times18\times1$\\\bottomrule

\end{tabular}

	}\vspace{1mm}%
	\caption{The network architecture for predictor module $\predictor$. The input to the network is obtained by concatenating the propagated states $\hat{h}_{t-1}$ ($18\times18\times8$), reliability score $\propconf_t$ ($18\times18\times1$), and appearance model output $s_t$ ($18\times18\times1$). All convolutional layers use $3\times3$ kernels. }
	\label{tab:predictor}%
	\vspace{-1mm}
\end{table}

\section{Detailed Results on VOT2018}
\label{sec:VOT2018}
\begin{figure}[t]
	\newcommand{\wid}{1.0\columnwidth}%
	\centering\vspace{0mm}%
	\includegraphics[trim = 0 0 0 0, width = \wid]{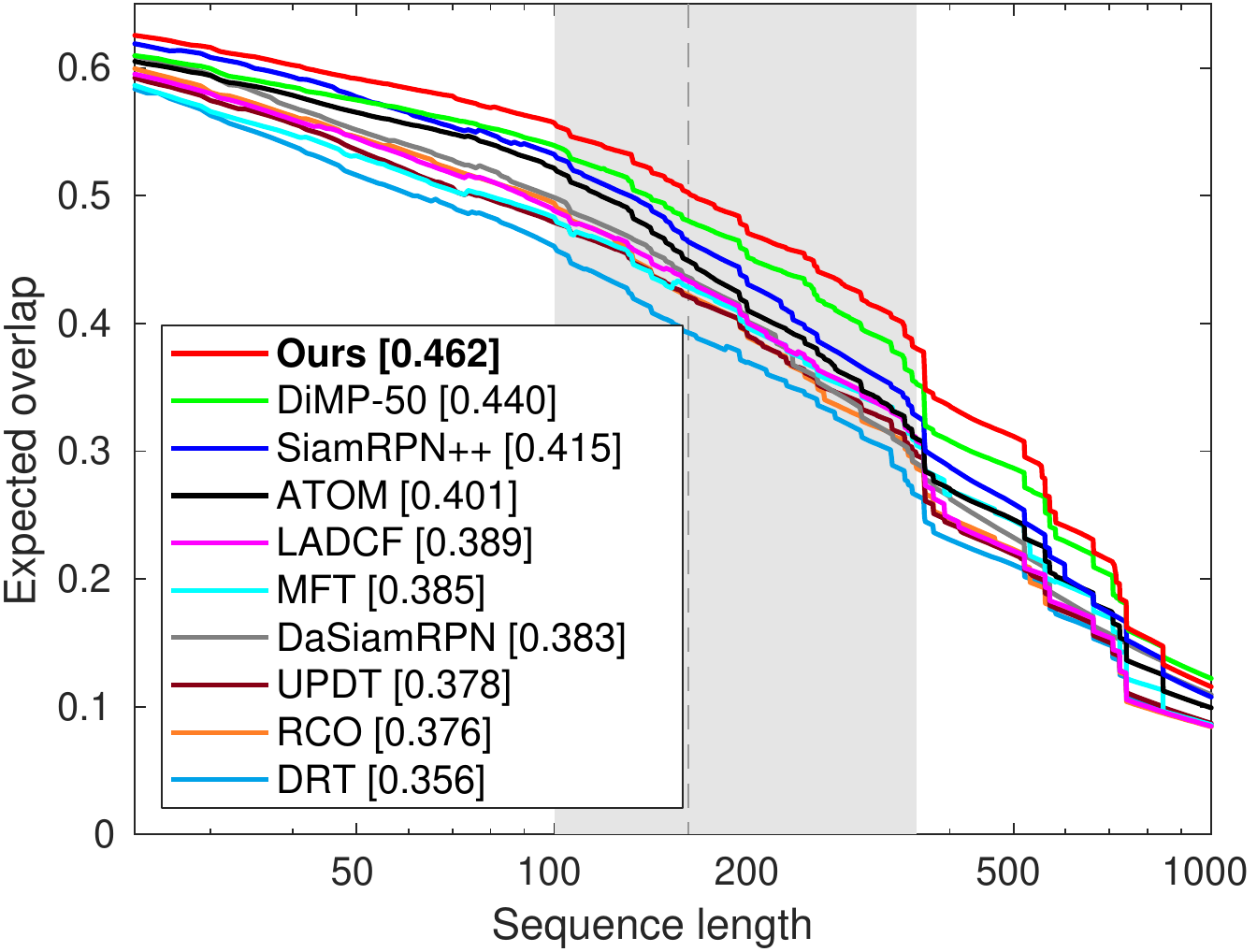}%
	\caption{Expected average overlap curve on the VOT2018 dataset. The plot shows the expected overlap between the tracker prediction and groundtruth for different sequence lengths. The expected average overlap (EAO) score, computed as the average of expected overlap values over typical sequence lengths (shaded region) is shown in the legend. Our tracker obtains the best EAO score, outperforming the previous best method DiMP-50 with a relative improvement of $5\%$ in EAO.}%
	\label{fig:vot}
\end{figure}

Here, we provide detailed results on the VOT2018 \cite{VOT2018} dataset, consisting of 60 challenging videos. The trackers are evaluated using the expected average overlap curve, which plots the expected average overlap between the tracker prediction and groundtruth for different sequence lengths. The average of the expected average overlap values over typical sequence lengths provides the expected average overlap (EAO) score, which is used to rank the trackers. We refer to \cite{VOT2015} for more details about EAO score computation.

We compare our approach with the recent state-of-the-art trackers: DRT~\cite{DRT}, RCO~\cite{VOT2018}, UPDT~\cite{BhatECCV2018}, DaSiamRPN~\cite{DaSiamRPN}, MFT~\cite{VOT2018}, LADCF~\cite{LADCF}, ATOM~\cite{ATOM}, SiamRPN++~\cite{SiamRPN++}, and DiMP-50~\cite{DiMP}. Figure~\ref{fig:vot} shows the expected average overlap curve. The EAO score for each tracker is shown in the legend. Our approach obtains the best results with an EAO score of $0.462$, outperforming the previous best method DiMP-50 with a relative improvement of $5\%$. This demonstrates the benefit of exploiting scene information for tracking.

\begin{figure*}[t]
	\centering%
	\newcommand{\wid}{0.24\textwidth}%
	\newcommand{\spc}{\hspace{1mm}}
	\includegraphics*[trim = 50 50 0 0, width = \wid]{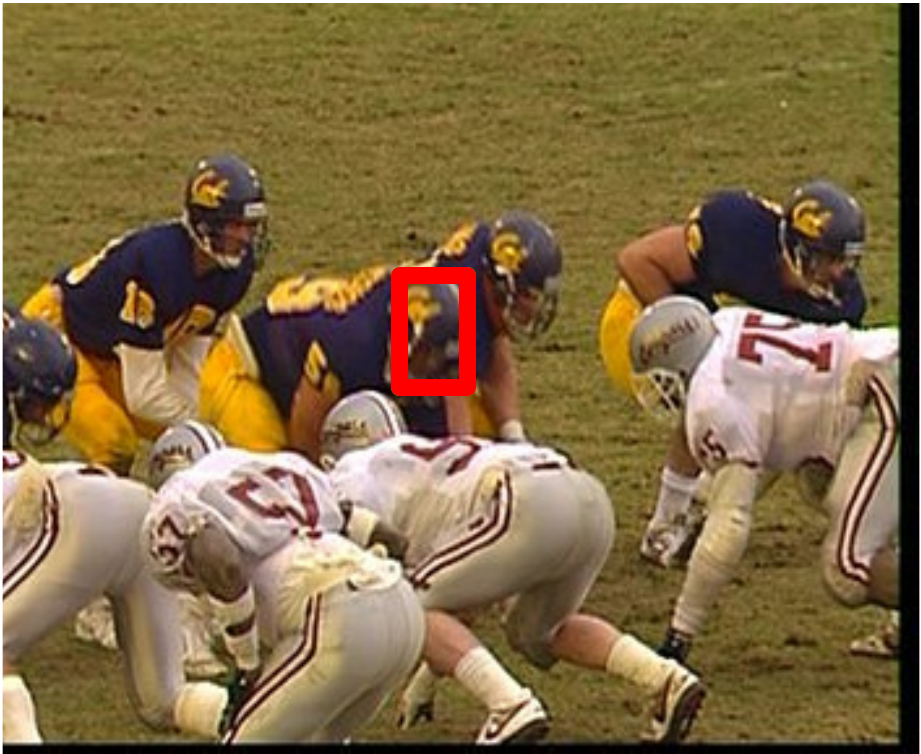}\spc%
	\includegraphics*[trim = 50 50 0 0, width = \wid]{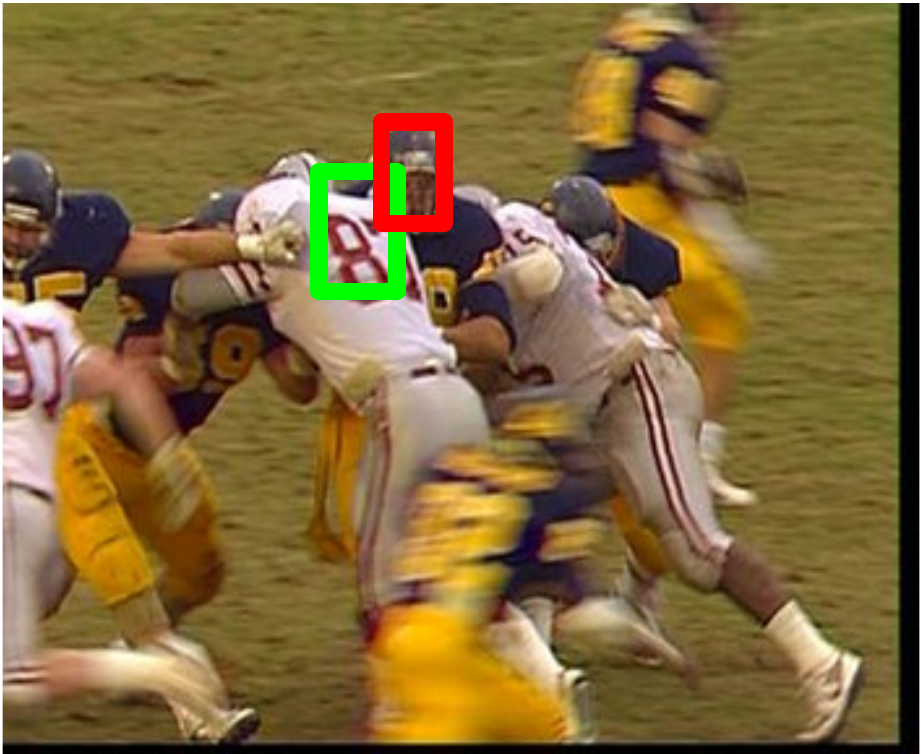}\spc%
	\includegraphics*[trim = 50 50 0 0, width = \wid]{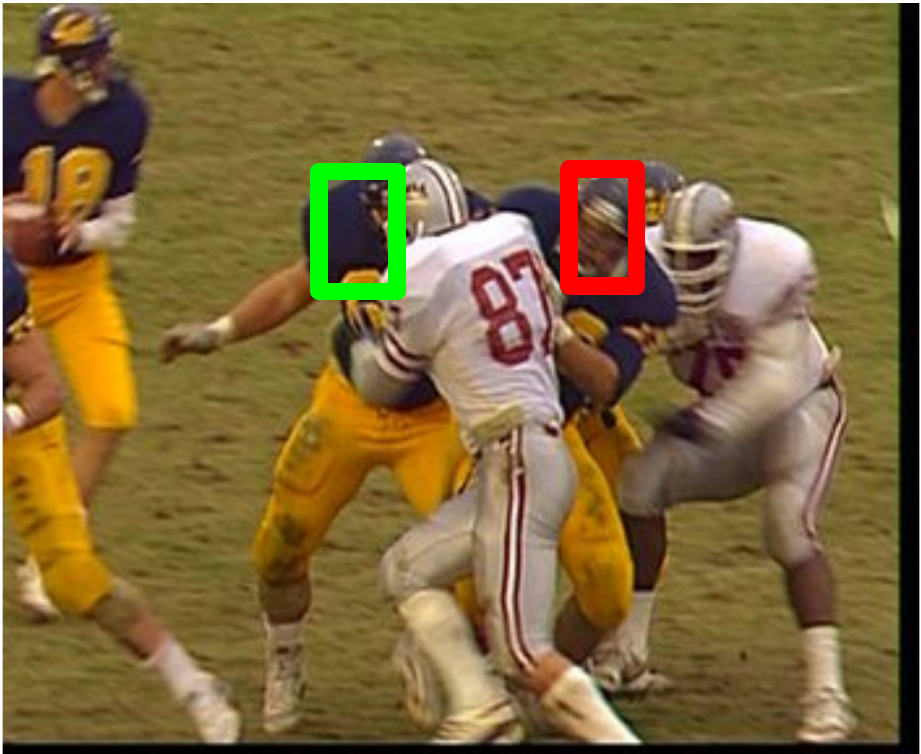}\spc%
	\includegraphics*[trim = 50 50 0 0, width = \wid]{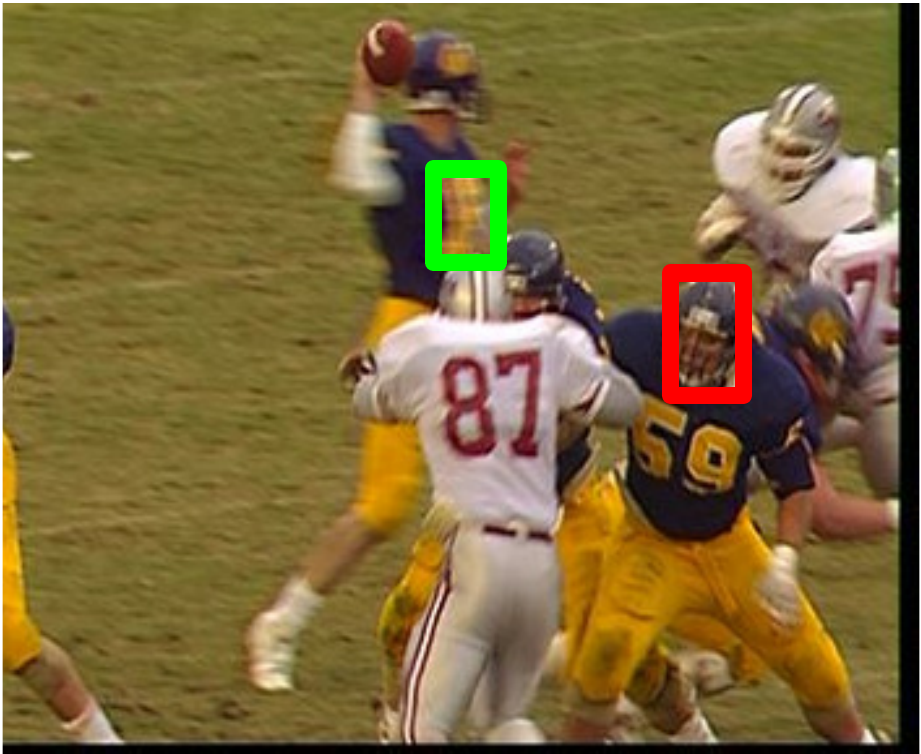}\vspace{2mm}
	\includegraphics*[trim = 0 280 0 100, width = \wid]{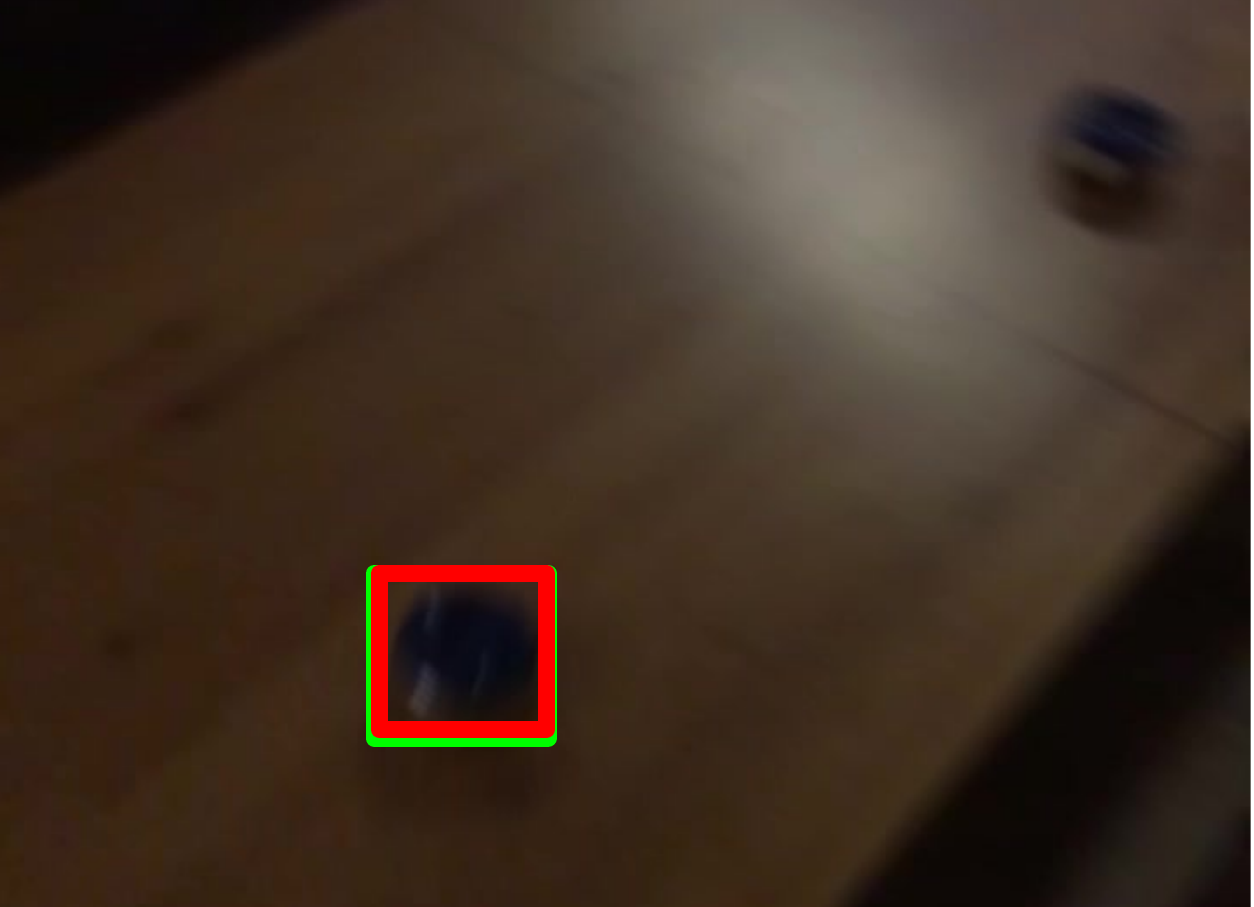}\spc%
	\includegraphics*[trim = 0 280 0 100, width = \wid]{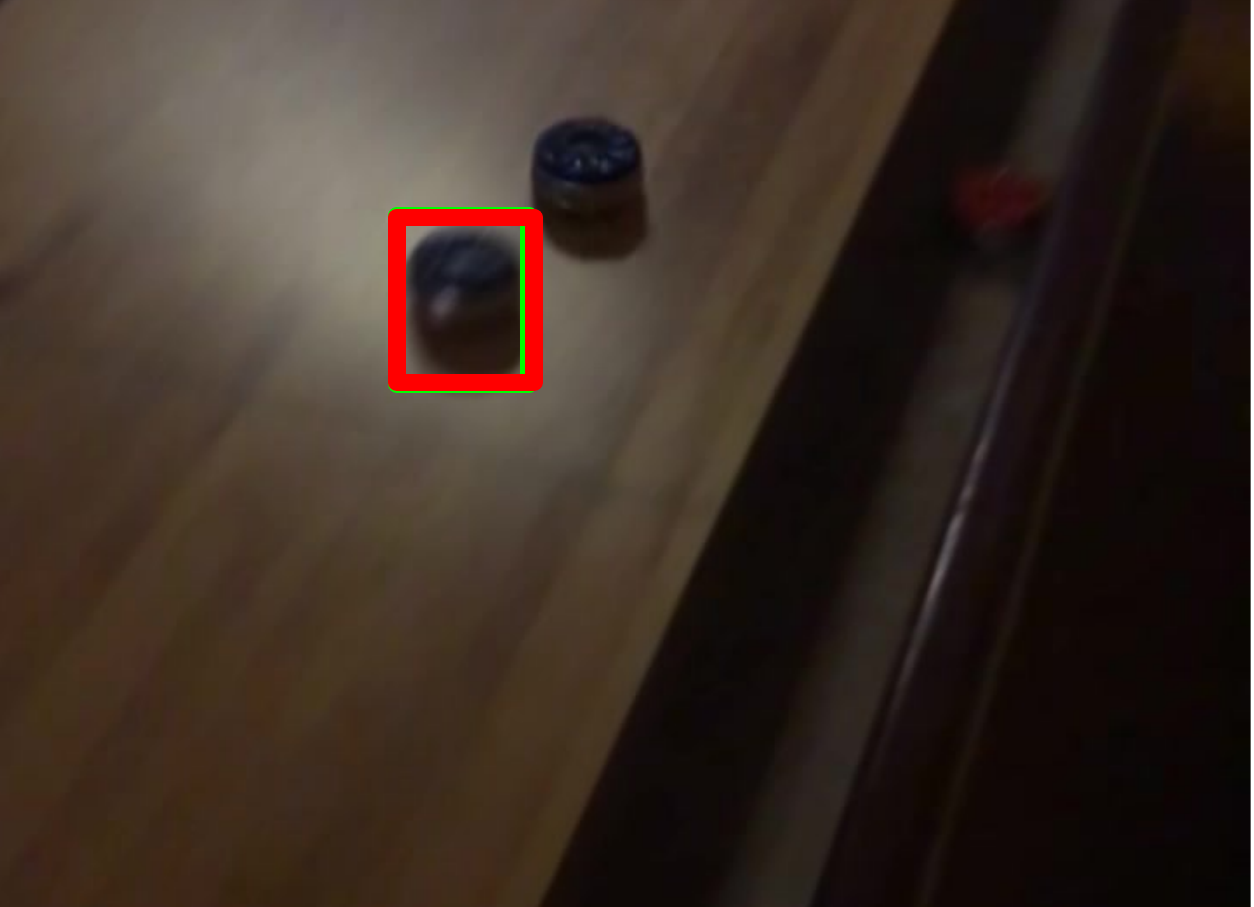}\spc%
	\includegraphics*[trim = 0 280 0 100, width = \wid]{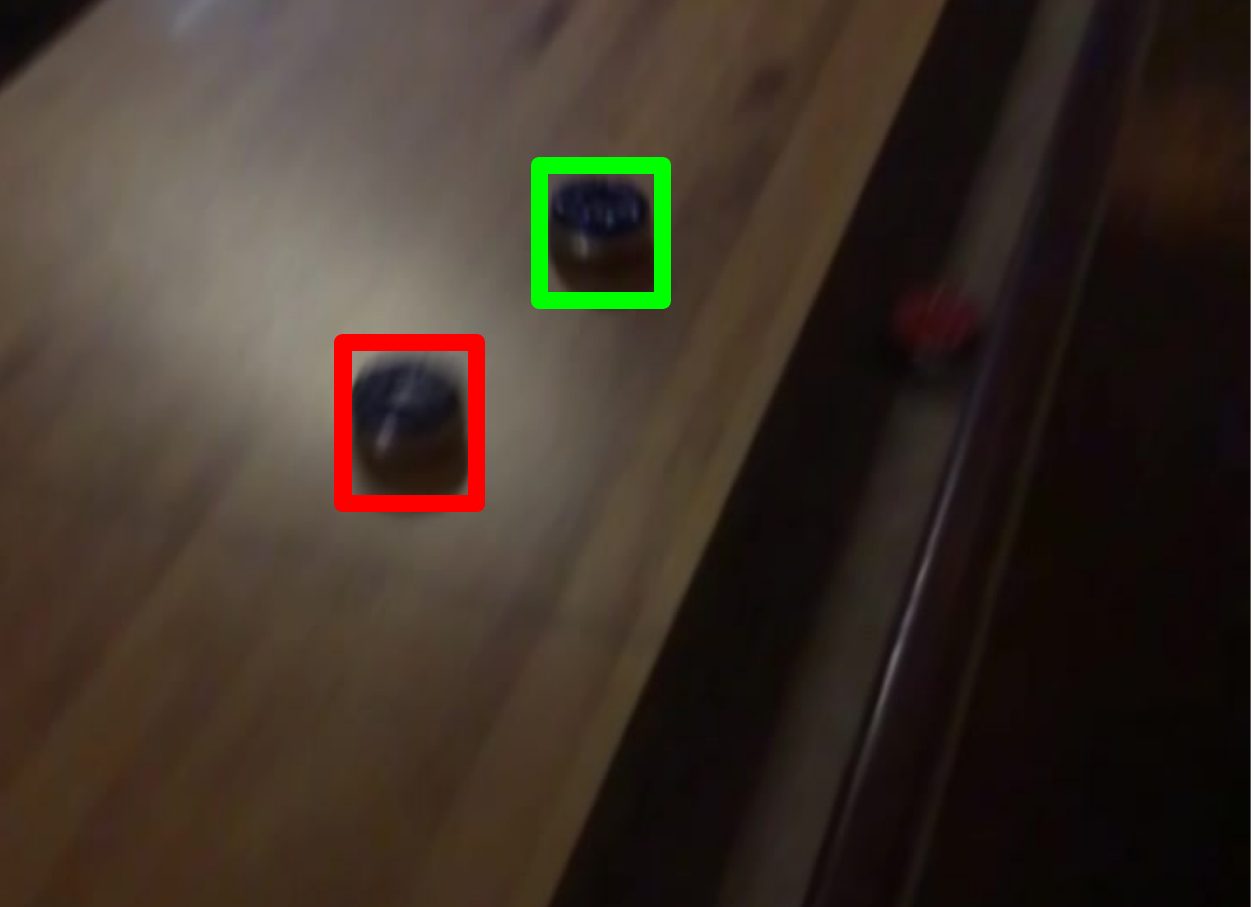}\spc%
	\includegraphics*[trim = 0 280 0 100, width = \wid]{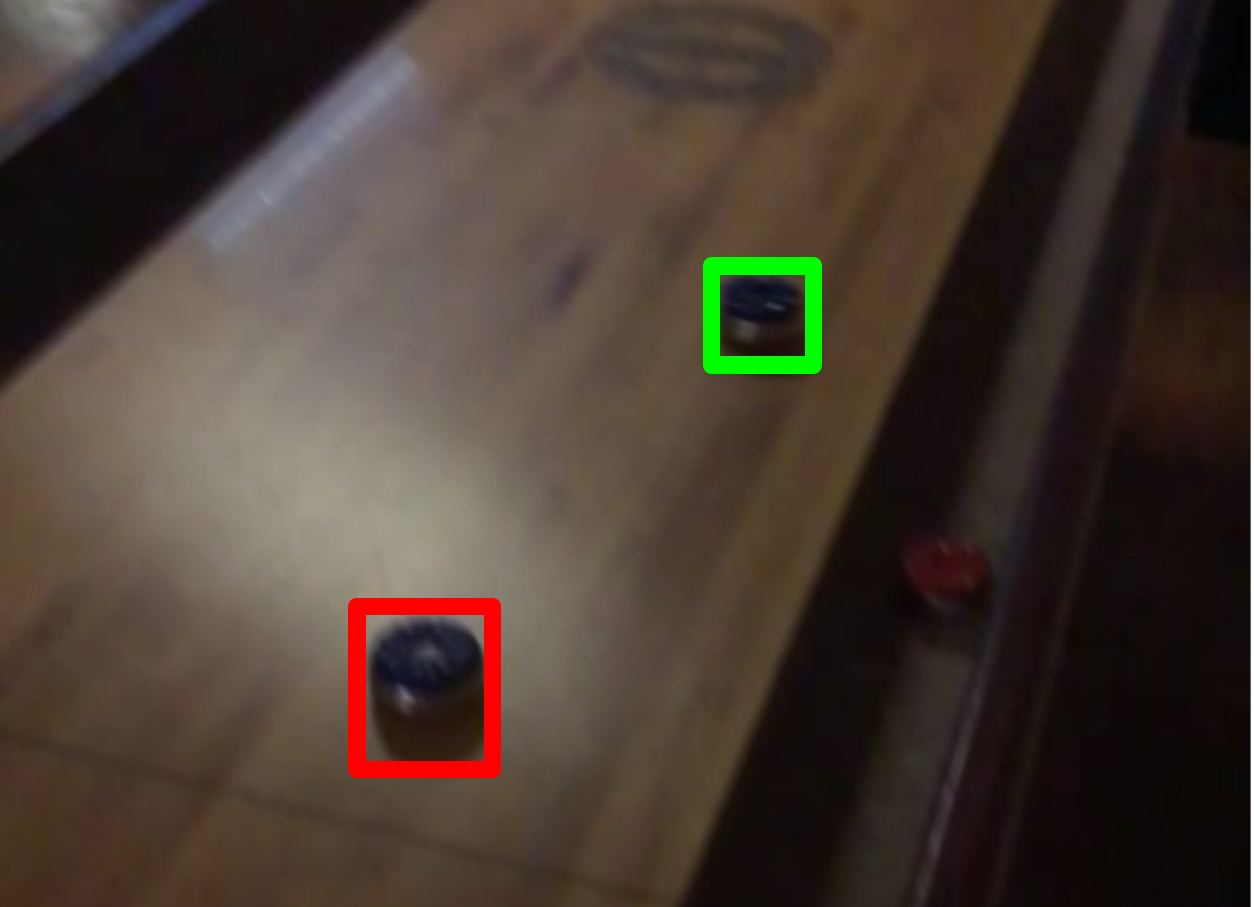}\vspace{2mm}
	\includegraphics*[trim = 300 250 280 0, width = \wid]{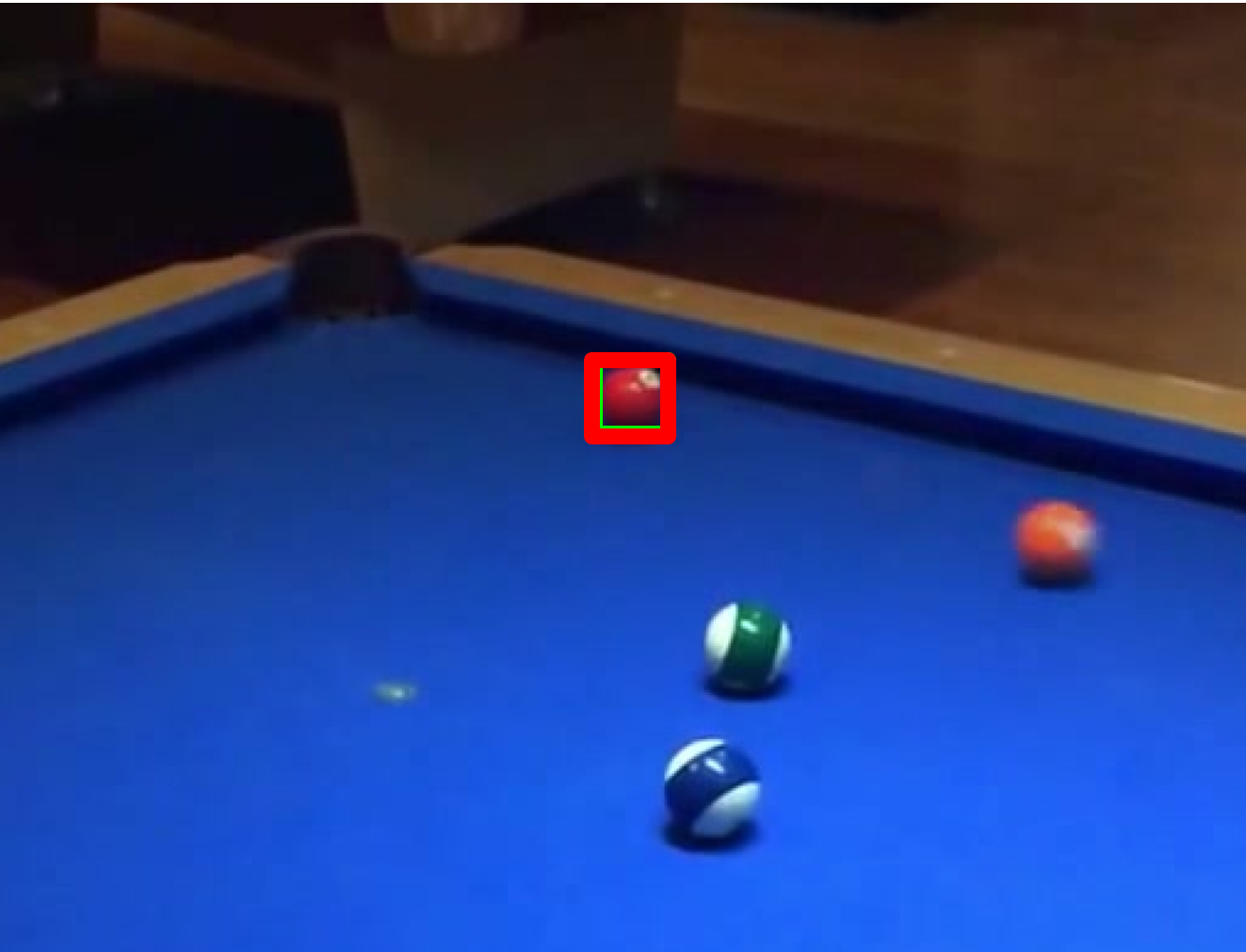}\spc%
	\includegraphics*[trim = 250 250 330 0, width = \wid]{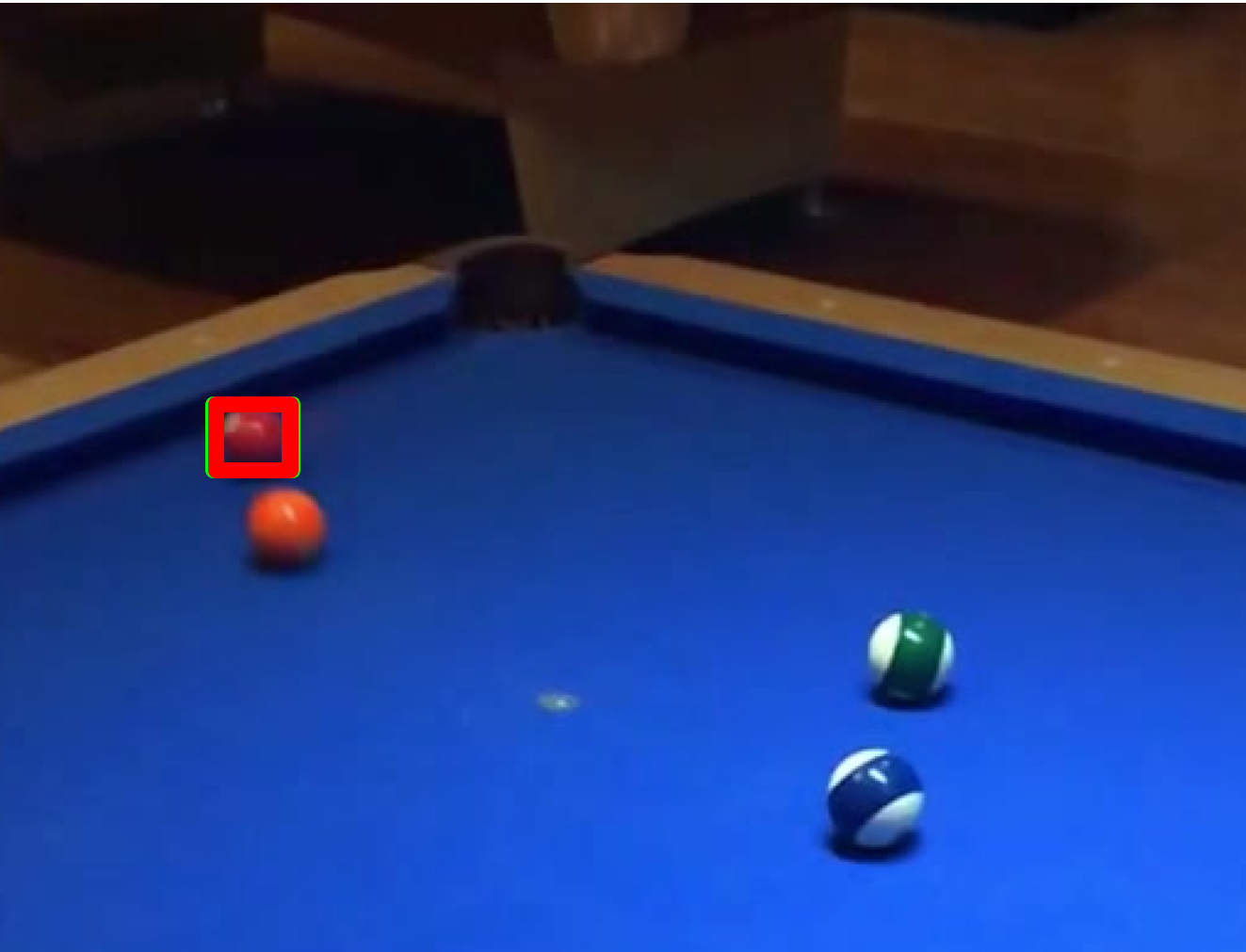}\spc%
	\includegraphics*[trim = 250 250 330 0, width = \wid]{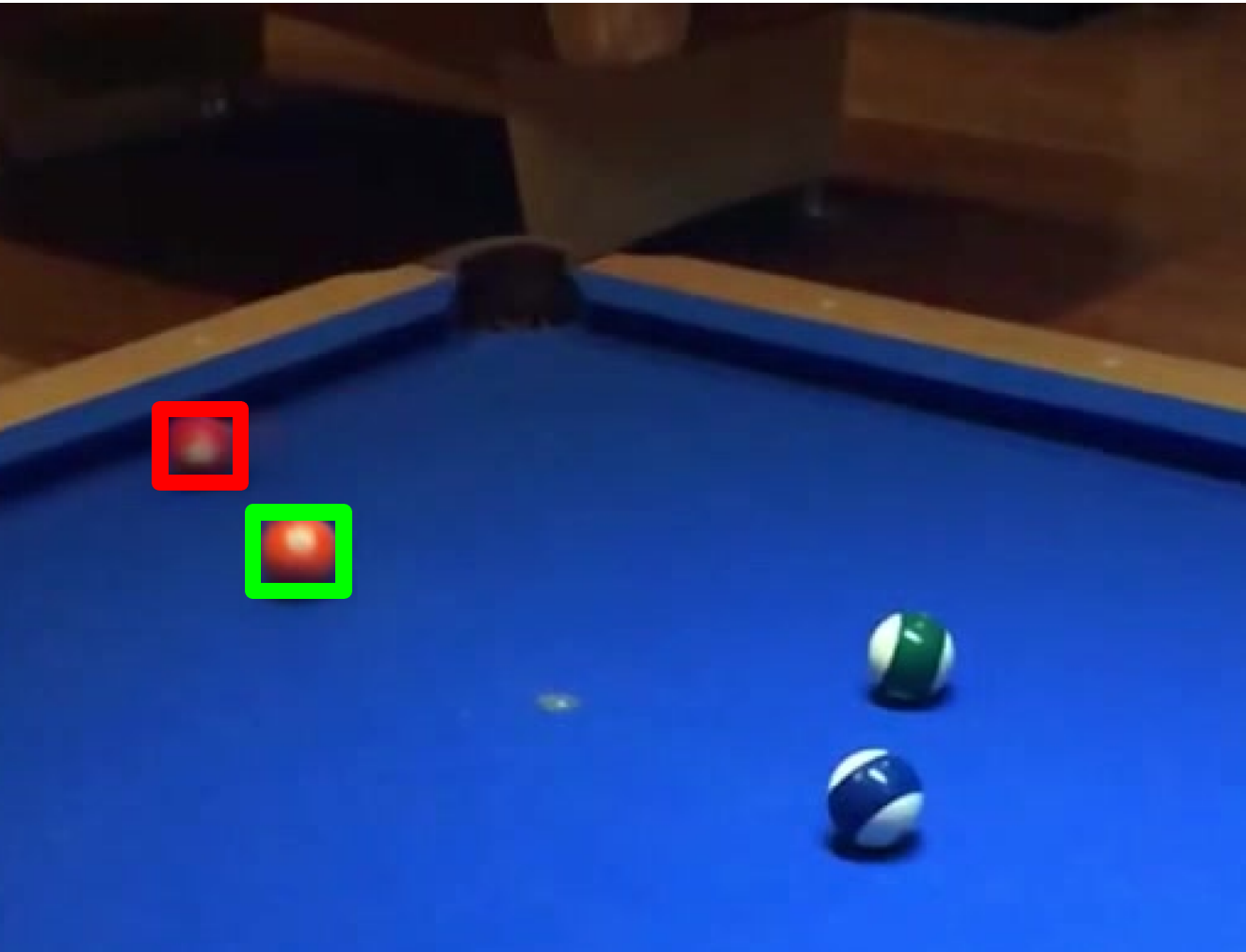}\spc%
	\includegraphics*[trim = 200 250 380 0, width = \wid]{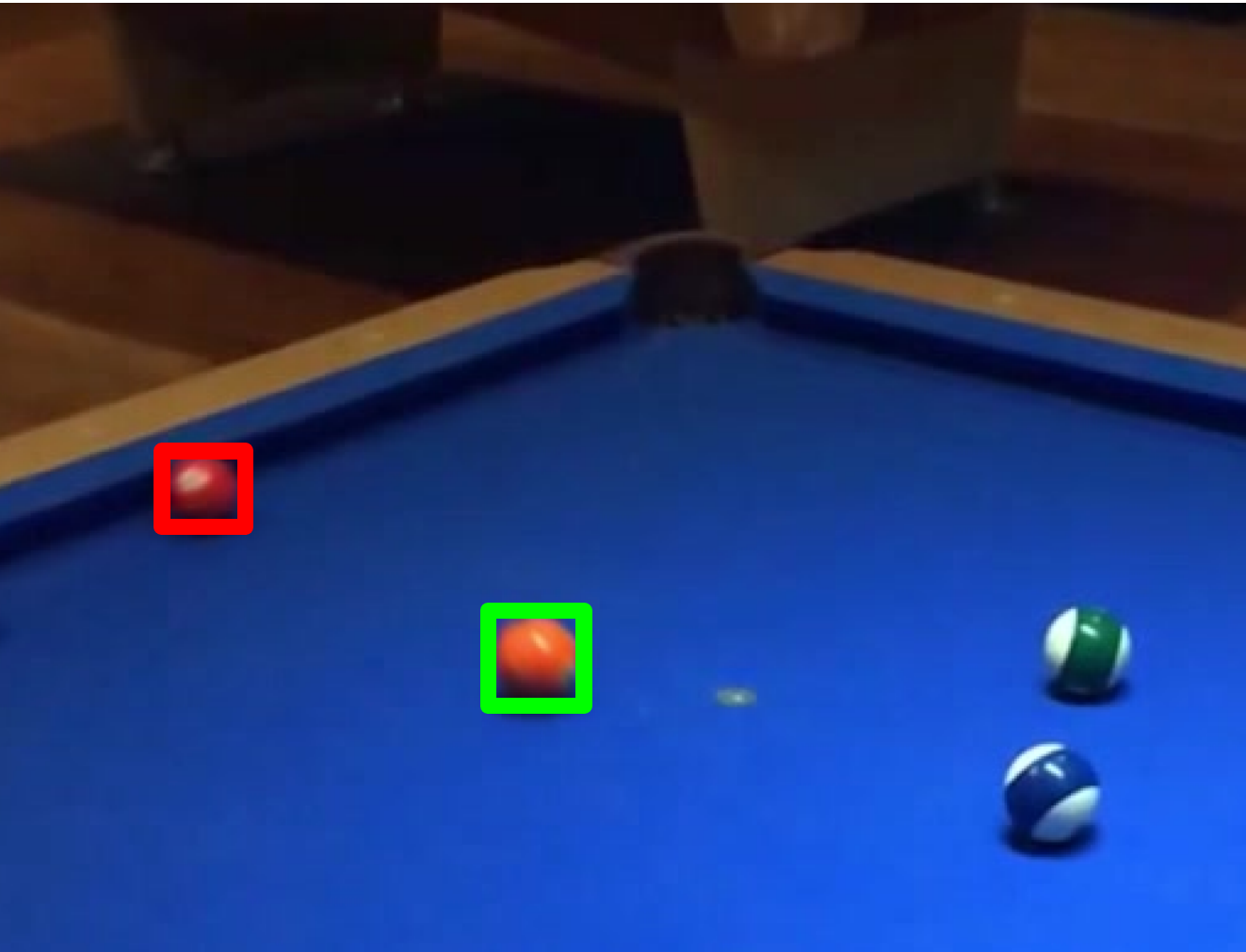}\vspace{2mm}
	\includegraphics*[trim = 0 30 250 0, width = \wid]{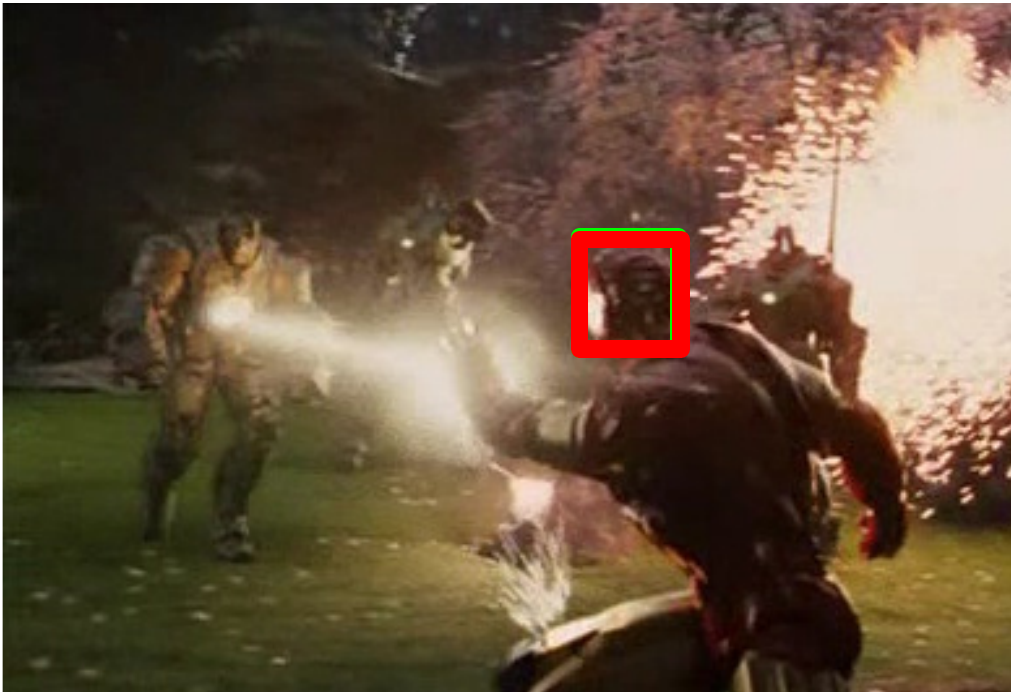}\spc%
	\includegraphics*[trim = 0 30 250 0, width = \wid]{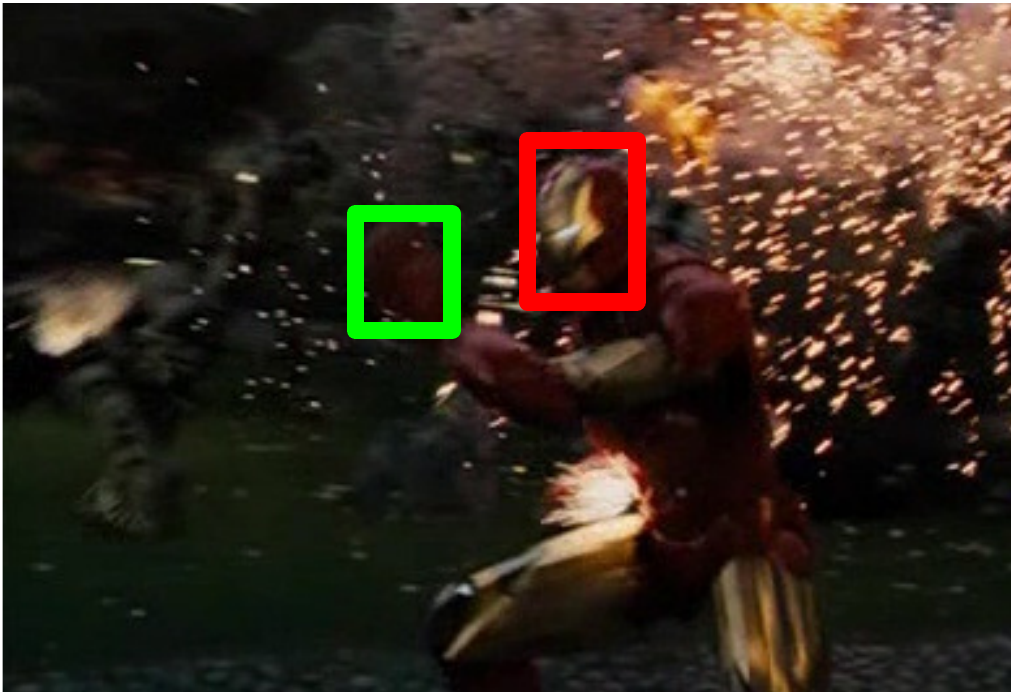}\spc%
	\includegraphics*[trim = 0 30 250 0, width = \wid]{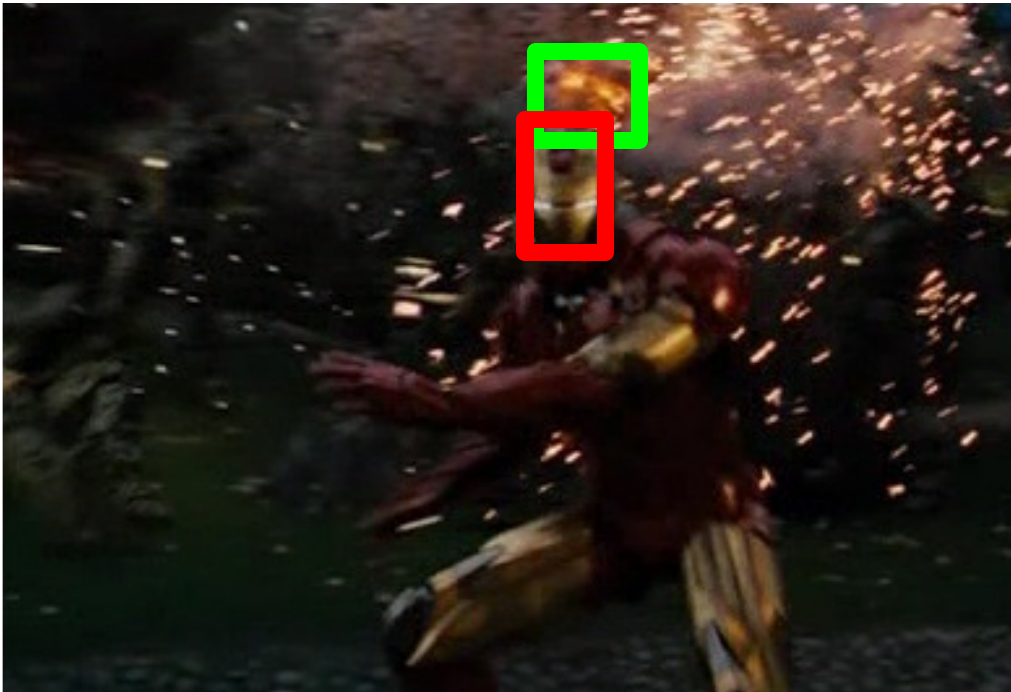}\spc%
	\includegraphics*[trim = 0 30 250 0, width = \wid]{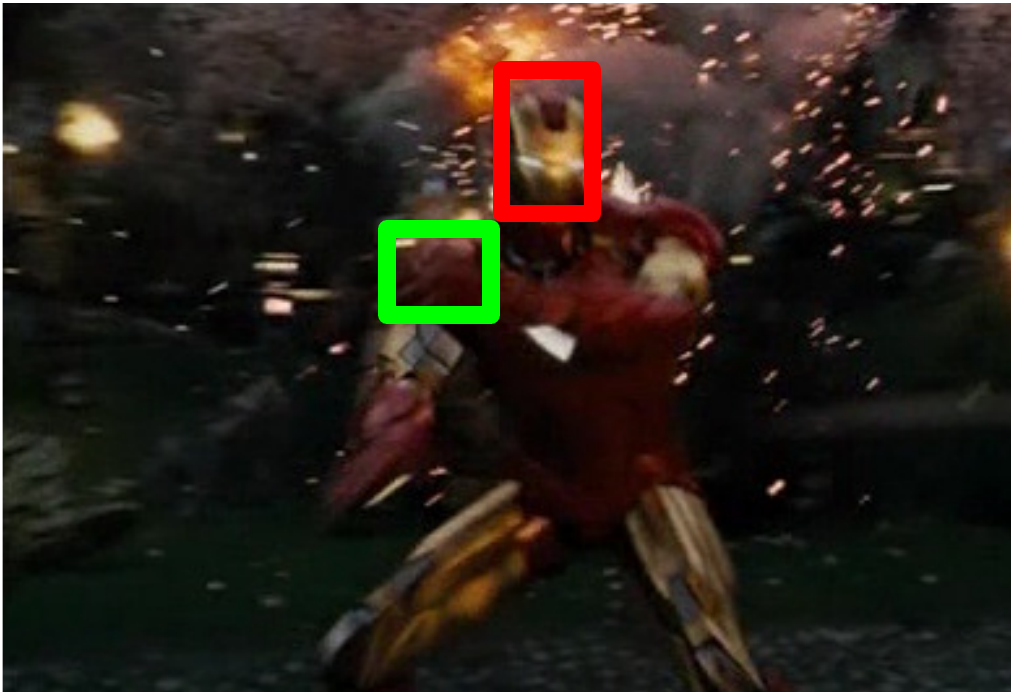}\vspace{2mm}
	\includegraphics*[trim = 0 200 0 150, width = \wid]{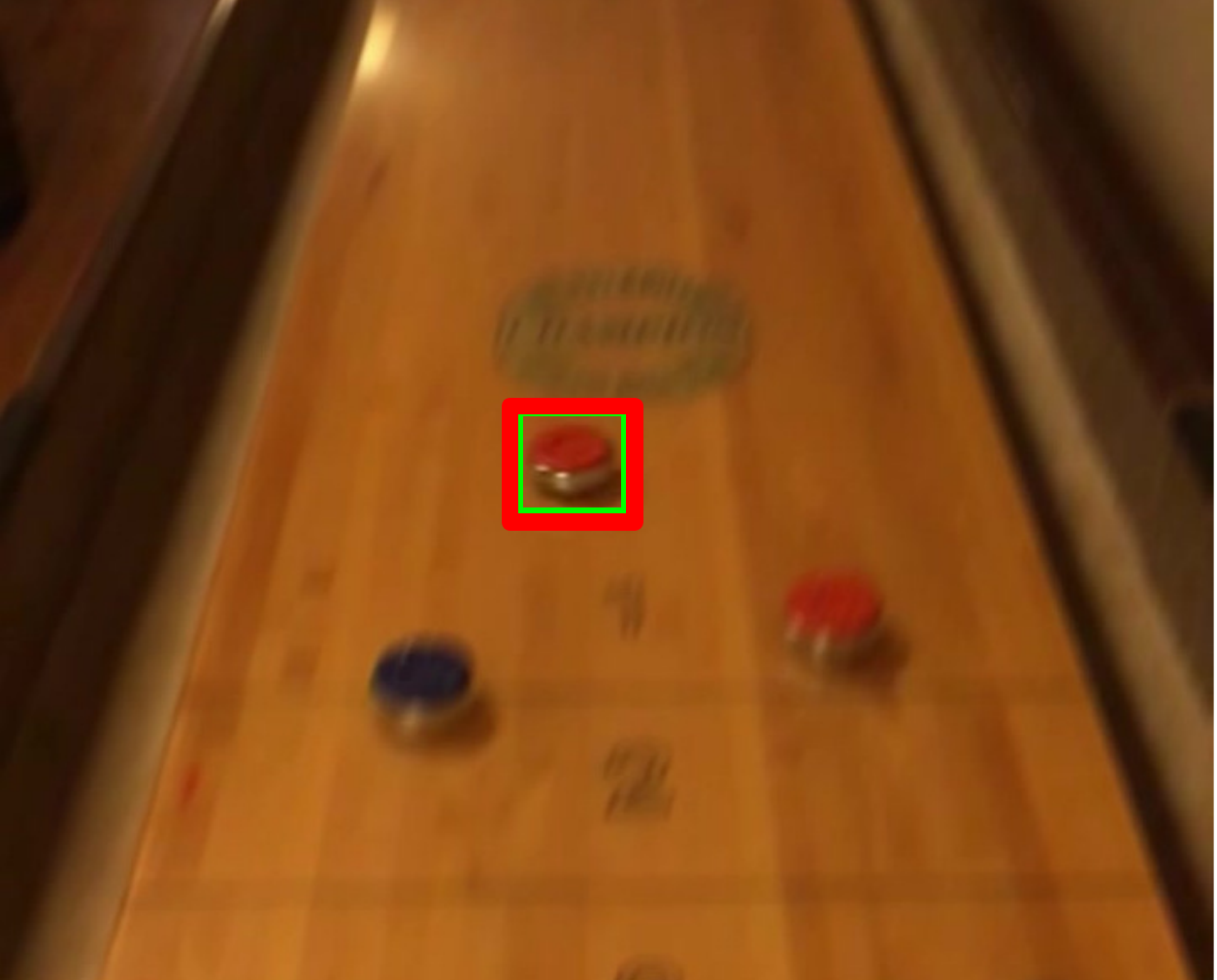}\spc%
	\includegraphics*[trim = 0 200 0 150, width = \wid]{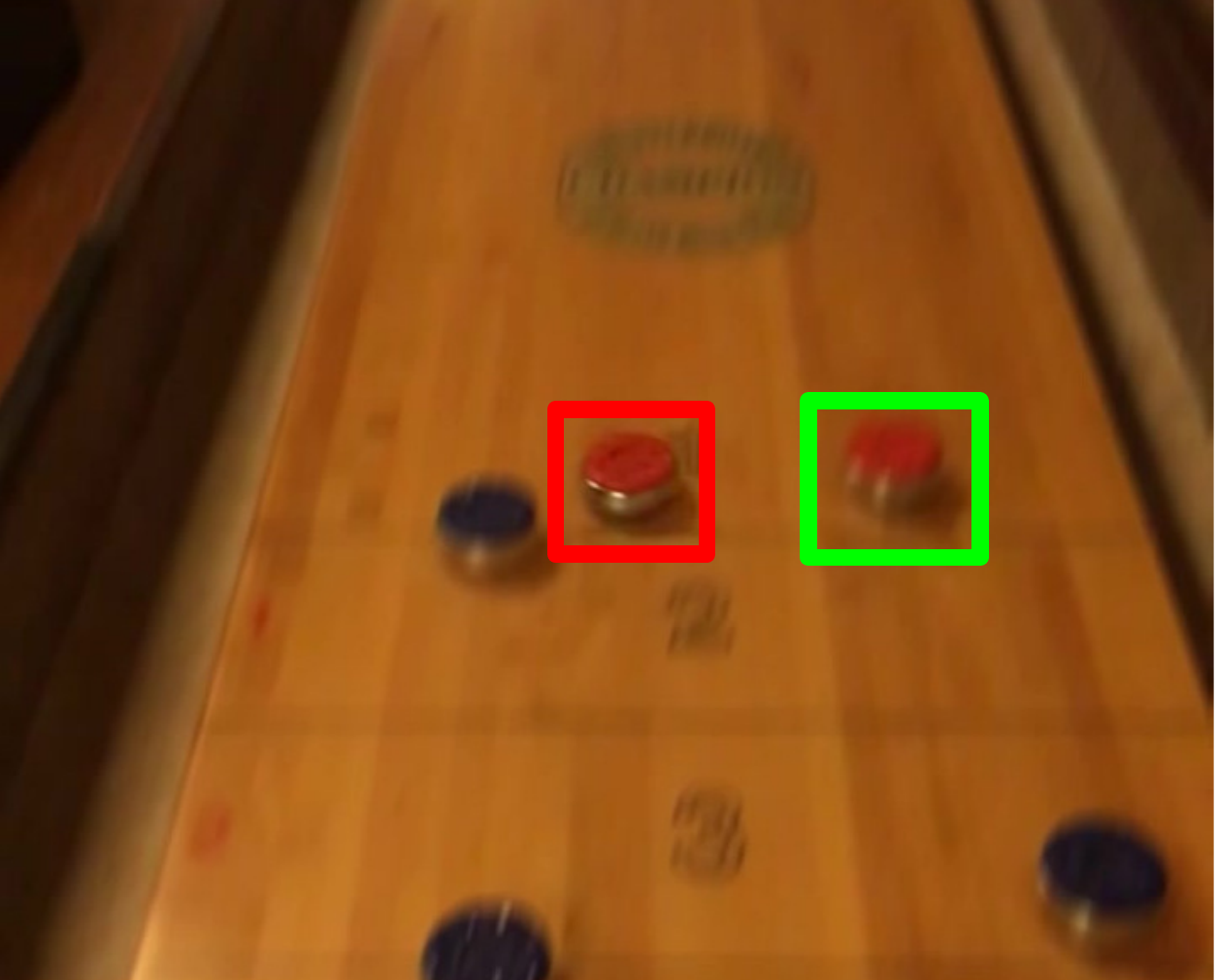}\spc%
	\includegraphics*[trim = 0 180 0 170, width = \wid]{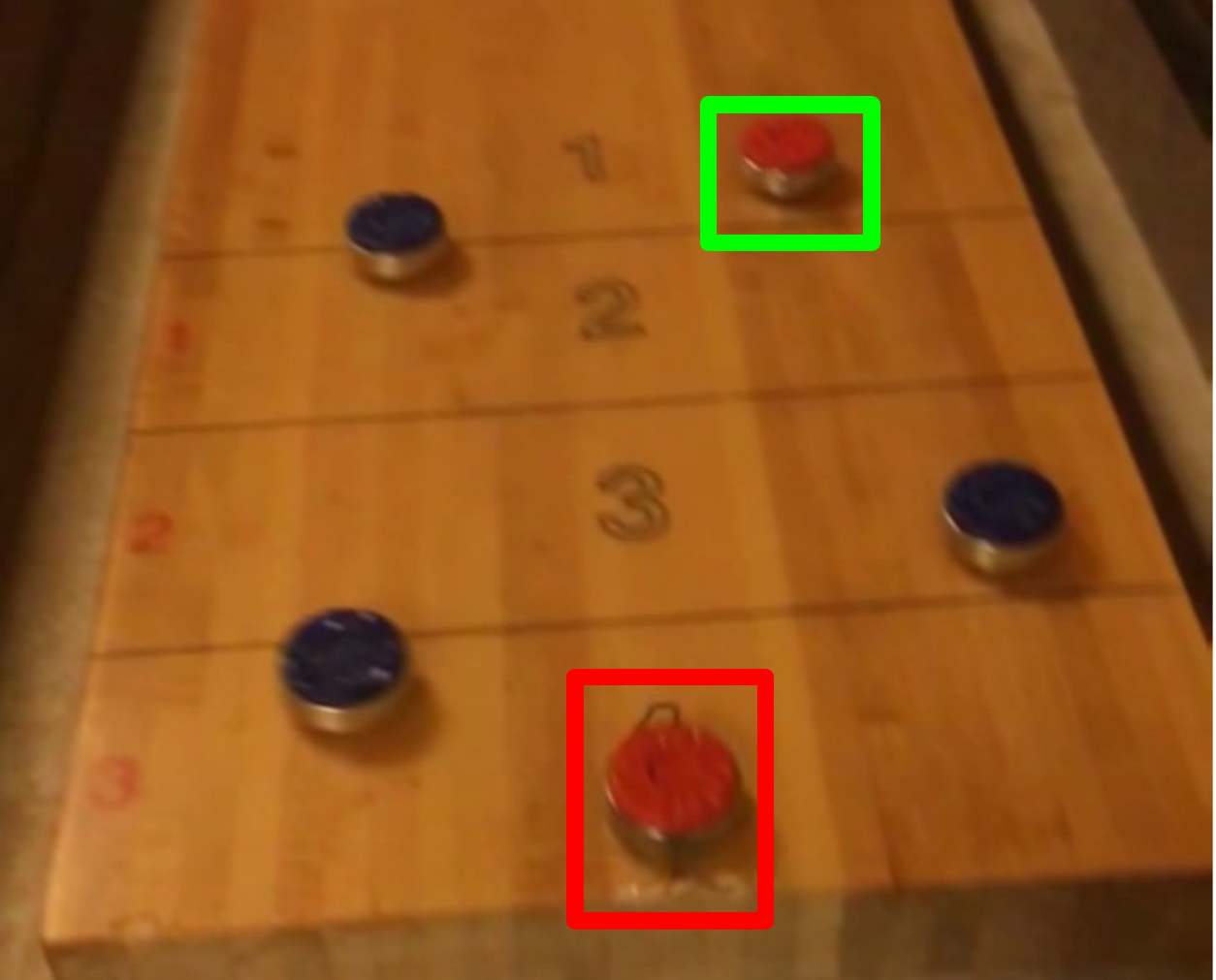}\spc%
	\includegraphics*[trim = 0 100 0 250, width = \wid]{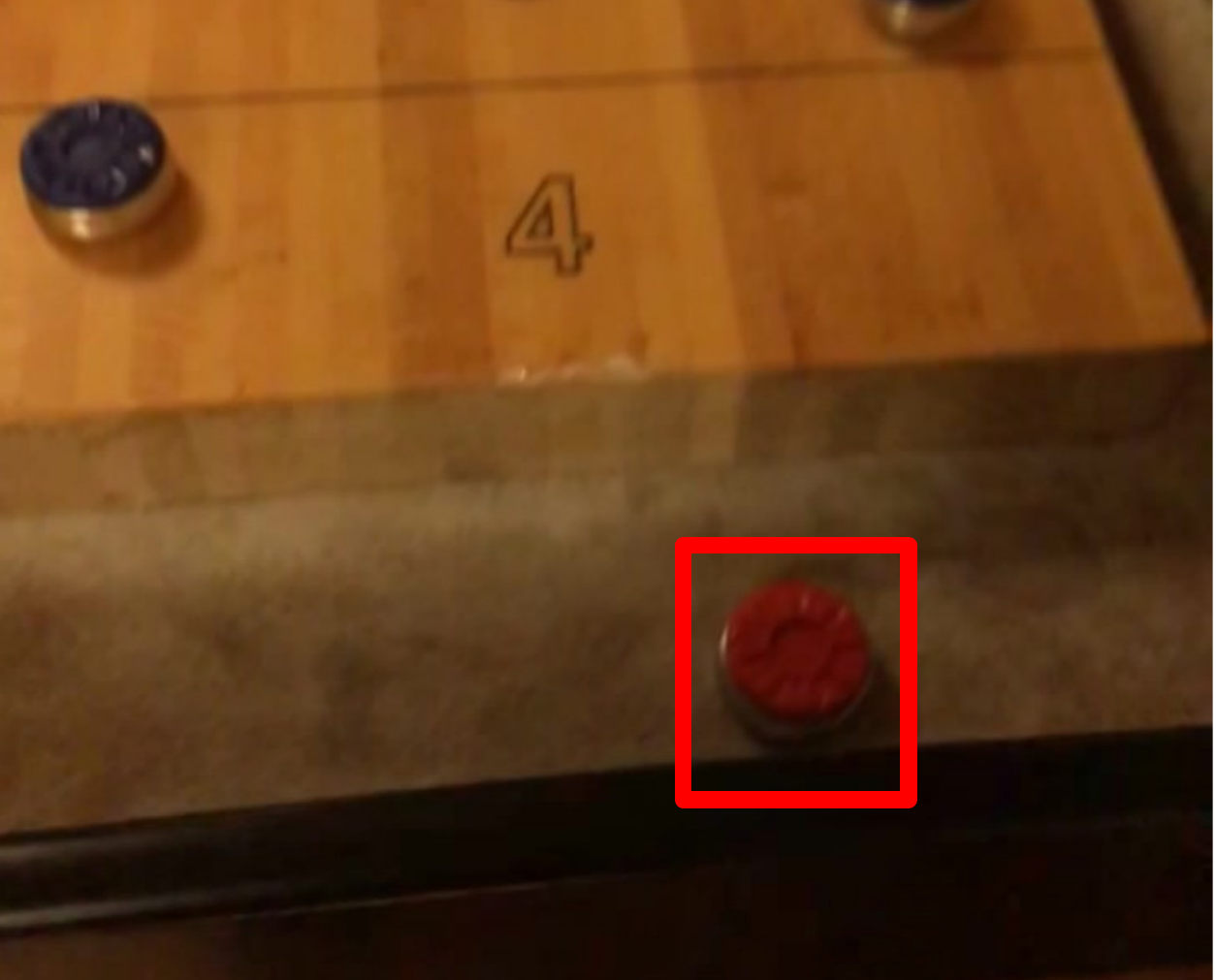}\vspace{2mm}
	\includegraphics*[trim = 400 300 350 100, width = \wid]{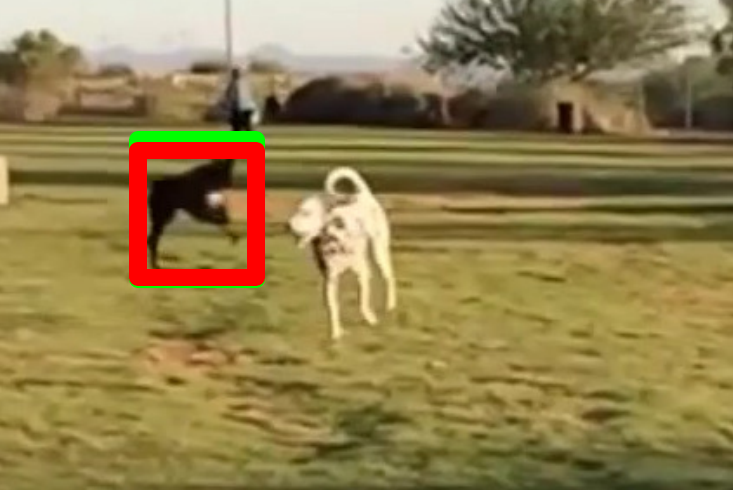}\spc%
	\includegraphics*[trim = 400 300 350 100, width = \wid]{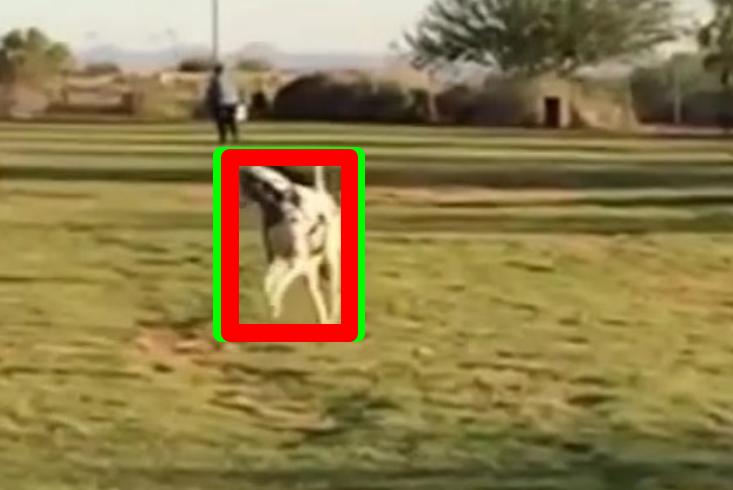}\spc%
	\includegraphics*[trim = 400 300 350 100, width = \wid]{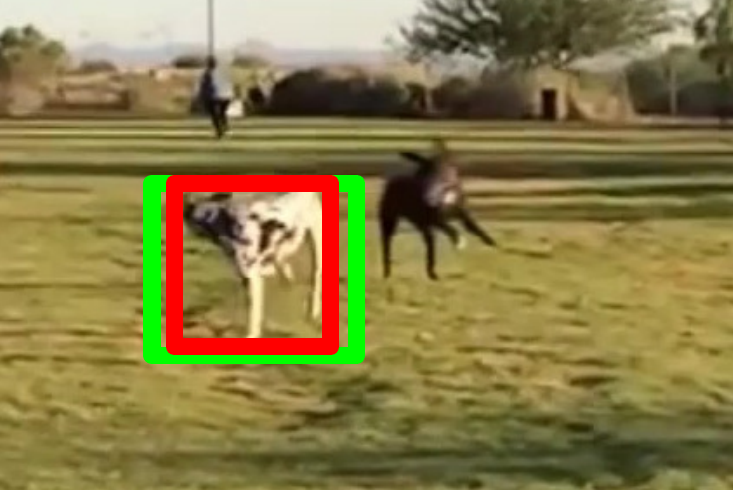}\spc%
	\includegraphics*[trim = 400 300 350 100, width = \wid]{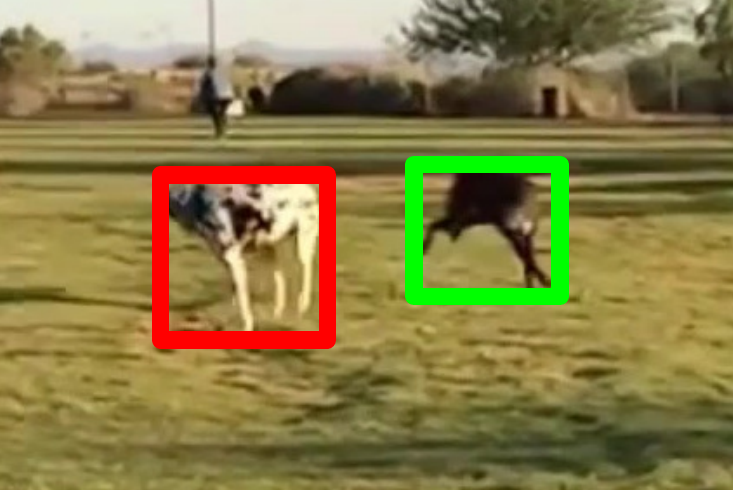}
	\includegraphics*[trim = 2 2 2 2, width = 0.65\columnwidth]{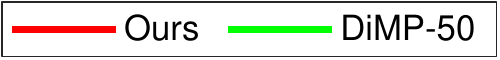}\vspace{0mm}%
	\caption{A qualitative comparison of our approach with the baseline appearance model, DiMP-50. Our tracker extracts information about other objects in the scene and exploits this knowledge to provide scene-aware predictions. Consequently, our approach can handle distractor objects which are hard to distinguish based on appearance only (second, third, and fifth rows). The propagated scene information is also beneficial to eliminate target candidate regions, which can be helpful in case of fast target appearance changes (first and fourth rows). The last row shows a failure case of our approach. Here, the appearance model cannot detect the occlusion caused by the white dog. This results in incorrect state updates, leading to tracking failure. 
	}\vspace{0mm}%
	\label{fig:qual}%
\end{figure*}

\section{Qualitative Results}
\label{sec:qual}
Here, we provide a qualitative comparison of our approach with the baseline tracker DiMP-50 \cite{DiMP}, which uses only an appearance model. Figure \ref{fig:qual} shows the tracking output for both the trackers on a few example sequences. DiMP-50 struggles to handle distractor objects which are hard to distinguish based on only appearance (second, third, fifth). In contrast, our approach is aware of the distractor objects in the scene and can exploit this scene information to achieve robust tracking. Propagating the scene information is also helpful in case of fast target appearance changes (first and fourth rows). In these cases, keeping track of the background regions can be useful to eliminate target candidate regions, greatly simplifying target localization. The last row shows a failure case of our approach. Here, the appearance model fails to detect the occlusion caused by the white dog. As a result, the state vectors are updated incorrectly, and the tracker starts tracking the white dog.

\end{document}